\documentclass[letterpaper, 10 pt, conference]{ieeeconf}  

\usepackage{amssymb}
\usepackage{pifont}
\usepackage{amsmath,bm}
\usepackage[T1]{fontenc}    
\usepackage{hyperref}       
\usepackage{url}            
\usepackage{booktabs}       
\usepackage{multirow}
\usepackage{amsfonts}       
\usepackage{nicefrac}       
\usepackage{microtype}      
\usepackage{subcaption}
\usepackage{caption}
\usepackage{graphicx}
\usepackage[ruled,vlined]{algorithm2e}
\usepackage{MnSymbol}
\usepackage{dsfont}

\usepackage{amsmath}
\usepackage{amssymb}
\usepackage{mathtools}
\usepackage{bm}
\usepackage{bbm}
\usepackage{comment}
\usepackage{makecell}
\usepackage{xcolor}

\usepackage{amssymb}
\usepackage{pifont}

\newtheorem{proposition}{Proposition}

\usepackage{comment}

\IEEEoverridecommandlockouts                              

\title{\LARGE \bf
Visually Robust Adversarial Imitation Learning from Videos with Contrastive Learning
}

\author{Vittorio Giammarino$^{1}$, James Queeney$^{2}$ and Ioannis Ch. Paschalidis$^{3}$
\thanks{VG and ICP were partially supported by the DOE under grants DE-AC02-05CH11231 and DE\-EE0009696, the NSF under grants CCF\-2200052, DMS\-1664644, and IIS\-1914792, and by the ONR under grants N00014\-19\-1\-2571 and N00014\-21\-1\-2844. JQ was exclusively supported by Mitsubishi Electric Research Laboratories (MERL).}
\thanks{$^{1}$Vittorio Giammarino is with Division of Systems Engineering, Boston University, Boston, MA 02215, USA
        {\tt\small vgiammar@bu.edu}}%
\thanks{$^{2}$James Queeney is with Mitsubishi Electric Research Laboratories (MERL), Cambridge, MA 02139, USA
        {\tt\small queeney@merl.com}}%
\thanks{$^{3}$Ioannis Ch. Paschalidis is with the Department of Electrical and Computer Engineering, Division of Systems Engineering, and Department of Biomedical Engineering, Boston University, Boston, MA 02215, USA
        {\tt\small yannisp@bu.edu}}%
}

\begin{document}

\maketitle
\thispagestyle{empty}
\pagestyle{empty}

\begin{abstract}
We propose C-LAIfO, a computationally efficient algorithm designed for imitation learning from videos in the presence of visual mismatch between agent and expert domains. We analyze the problem of imitation from expert videos with visual discrepancies, and introduce a solution for robust latent space estimation using contrastive learning and data augmentation. Provided a visually robust latent space, our algorithm performs imitation entirely within this space using off-policy adversarial imitation learning. We conduct a thorough ablation study to justify our design and test C-LAIfO on high-dimensional continuous robotic tasks. Additionally, we demonstrate how C-LAIfO can be combined with other reward signals to facilitate learning on a set of challenging hand manipulation tasks with sparse rewards. Our experiments show improved performance compared to baseline methods, highlighting the effectiveness of C-LAIfO. To ensure reproducibility, we open source our \href{https://github.com/VittorioGiammarino/C-LAIfO}{code}.
\end{abstract}

\section{INTRODUCTION}

\begin{figure}[h!]
    \centering
    \begin{subfigure}[t]{0.32\linewidth}
        \centering
        \includegraphics[width=\linewidth]{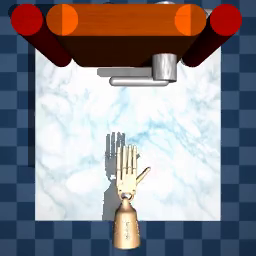}
        \caption{}
        \label{fig:door_intro}
    \end{subfigure}
    \begin{subfigure}[t]{0.32\linewidth}
        \centering
        \includegraphics[width=\linewidth]{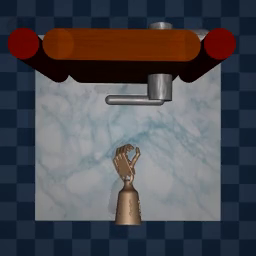}
        \caption{}
        \label{fig:door-light_intro}
    \end{subfigure}
    \begin{subfigure}[t]{0.32\linewidth}
        \centering
        \includegraphics[width=\linewidth]{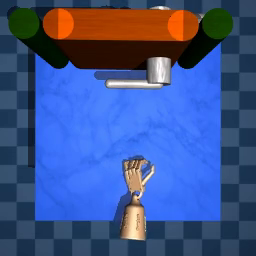}
        \caption{}
        \label{fig:door-color_intro}
    \end{subfigure}
    \caption{Robotic manipulation task. Current end-to-end methods for imitation from expert videos assume that the expert and the agent operate in the same environment. Consequently, they are unable to handle variations in lighting or background.}
    \label{fig:adroit_intro}
\end{figure}

In recent years, there has been a significant surge in research on imitation learning from expert videos, commonly referred to as the \textit{Visual Imitation from Observations (V-IfO)} problem. The approach of mimicking experts from videos holds great promise for the future, as it offers a cost-effective way to teach autonomous agents new skills and behaviors. To achieve this goal, prior research has developed methods capable of concurrently addressing two primary challenges of the V-IfO framework: the partial observability of the decision-making process and the absence of expert actions \cite{giammarino2023adversarial}. Despite these advancements, end-to-end state-of-the-art algorithms still face significant barriers in real-world applications due to the assumption that both the expert and learning agent operate within the same environment \cite{giammarino2023adversarial, liu2022visual}. For instance, consider the scenario described in Fig.~\ref{fig:adroit_intro}, where expert videos are collected under the conditions in Fig.~\ref{fig:door_intro} and an autonomous agent is deployed in Fig.~\ref{fig:door-light_intro} or Fig.~\ref{fig:door-color_intro}. Current methods are not designed to handle such variations in lighting and background, leading to failures in these contexts. Our goal, in this paper, is to enhance the imitation capabilities of autonomous agents in the presence of visual mismatches.   

We introduce a novel end-to-end pipeline for imitation from expert videos with visual mismatch. We begin by analyzing the V-IfO problem with visual mismatch and propose a novel, simple, and computationally efficient algorithm called Contrastive Latent Adversarial Imitation from Observations (C-LAIfO). Notably, C-LAIfO builds upon the recent LAIfO algorithm \cite{giammarino2023adversarial} and achieves visually robust latent state estimation through data augmentation and contrastive learning techniques \cite{chen2020simple, grill2020bootstrap}. We justify each design choice for our algorithm, including the types of data augmentation and contrastive loss, through a comprehensive ablation study. Furthermore, we compare C-LAIfO against two V-IfO baselines—LAIfO \cite{giammarino2023adversarial} and PatchAIL \cite{liu2022visual}—as well as DisentanGAIL \cite{cetin2020domain}, which serves as a baseline for V-IfO with domain mismatch. Additionally, we show how the reward signal learned from expert data using C-LAIfO can be easily integrated with other signals to enhance efficiency and enable learning in robotic tasks with sparse reward functions. Therefore, we further evaluate our algorithm on the Adroit platform for dynamic dexterous manipulation \cite{Kumar2016thesis}. These additional experiments highlight the versatility of our approach, showcasing its efficacy in handling complex robotic tasks.

\section{Related Work}
\label{sec:related_work}
\paragraph{Imitation from observation} \textit{Imitation Learning (IL)} is a powerful approach that allows agents to mimic expert behavior by using demonstrations of a task typically in the form of state-action pairs. Our work builds on \textit{Adversarial Imitation Learning (AIL)} \cite{ho2016generative, fu2017learning} which frames IL as a two-player game between a discriminator and an agent's policy. Here, the discriminator distinguishes whether a state-action pair is generated by the agent or the expert policy. In practice, AIL is formulated as a joint process of \textit{Reinforcement Learning (RL)} and inverse RL \cite{russell1998learning, ng2000algorithms, abbeel2004apprenticeship, ziebart2008maximum}. First a reward function is inferred from expert demonstrations and then it is used in the RL step to train agents. In scenarios with partial observability, AIL has been applied to cases with missing information \cite{gangwani2020learning} and to Visual IL, where agents learn from video frames as state observations \cite{rafailov2021visual}. Compared to standard IL, Imitation from observation \cite{torabi2018generative, yang2019imitation, cheng2021guaranteed} assumes that action information is not observable in the demonstrations data. This setting is more practical compared to IL but also more difficult to tackle. The combination of learning from videos in the absence of expert actions gives rise to the V-IfO problem, which is the primary focus of our work. End-to-end state-of-the-art algorithms for the V-IfO setting include PatchAIL \cite{liu2022visual}, which applies AIL directly on the pixel space utilizing a PatchGAN discriminator \cite{isola2017image, zhu2017unpaired}, and LAIfO \cite{giammarino2023adversarial}, where AIL operates on a latent representation of the agent state. Notably, these approaches are built on the assumption that both the expert and the learning agent act within the same decision process, which rarely holds true in real-world scenarios. 

\paragraph{Imitation from videos with environment mismatch} Our research targets imitation from visual observations in the presence of mismatches between the expert and the learner environments, a problem referred to as third-person IL \cite{stadie2017third}, domain-adaptive IL \cite{kim2020domain}, or cross-domain IL \cite{raychaudhuri2021cross}. Solutions in the literature either decompose this problem into sequential stages or formulate end-to-end methods. Sequential approaches include: \cite{liu2018imitation}, where a reward function is learned by leveraging video prediction with context translation; \cite{sermanet2018time}, where reward functions are obtained using time-contrastive networks trained offline; \cite{smith2020avid}, where cycle-consistent adversarial networks \cite{zhu2017unpaired} are trained offline to generate instruction images in the agent domain from videos; \cite{giammarino2023opportunities}, which uses inverse models and adversarial domain adaptation \cite{tzeng2017adversarial} to train navigation policies from videos, and \cite{zhang2023slomo}, where 3D trajectory reconstruction from videos is used to obtain physically plausible trajectories. Our work differs from this literature as we formulate a fully end-to-end approach. 

\paragraph{End-to-end algorithms for imitation from videos with mismatch} Previous end-to-end solutions were presented in \cite{stadie2017third, okumura2020domain, cetin2020domain, choi2023domain}. The studies in \cite{stadie2017third} and \cite{cetin2020domain} extract domain-independent features to infer domain-independent reward functions. More specifically, in \cite{stadie2017third}, the authors propose to learn domain-independent features using an adversarial approach similar to \cite{ganin2015unsupervised}, while DisentanGAIL in \cite{cetin2020domain} achieves a similar result by adding a mutual information constraint to the binary cross-entropy loss used for AIL. Similar to our algorithm, these studies formulate fully end-to-end model-free algorithms that avoid costly generative steps during the imitation process. Our approach adopts similar reasoning to \cite{stadie2017third, cetin2020domain}; however, we leverage contrastive learning for domain-independent feature extraction and build the entire AIL pipeline (both reward inference and RL step) on this learned feature space, rather than only the reward inference as done in \cite{stadie2017third, cetin2020domain}. As shown in our experiments, this leads to significant improvements in performance. Other works rely on expensive generative steps to address the mismatch problem. In \cite{okumura2020domain}, imitation is performed using expert observation-action pairs through a learned domain-agnostic recurrent state space model \cite{hafner2019learning}. Our algorithm, on the other hand, is model-free and only requires expert observations. In \cite{choi2023domain}, cycle-consistent adversarial networks \cite{zhu2017unpaired} are trained online to generate expert videos in the agent's domain, thus reducing the problem to the standard V-IfO without mismatches. Our approach does not require such a generative step, as it learns a domain-independent feature space directly.

\vspace{-0.6cm}
\section{Preliminaries}
\label{sec:preliminaries}
We use uppercase letters (e.g., $S_t$) for random variables, lowercase letters (e.g., $s_t$) for values of random variables, script letters (e.g., $\mathcal{S}$) for sets, and bold lowercase letters (e.g., $\bm{\theta}$) for vectors. Let $[t_1 : t_2]$ be the set of integers $t$ such that $t_1 \leq t \leq t_2$; we write $S_t$ such that $t_1 \leq t \leq t_2$ as $S_{t_1 : t_2}$. We denote with $\mathbb{E}[\cdot]$ expectation, with $\mathbb{P}(\cdot)$ probability, and with $\mathbb{D}_f(\cdot, \cdot)$ an $f$-divergence between two distributions of which the Jensen-Shannon divergence, $\mathbb{D}_{\text{JS}}(\cdot||\cdot)$, is a special case.

\paragraph{Partially Observable Markov Decision Process} We model the decision processes as infinite-horizon discounted Partially Observable Markov Decision Processes (POMDPs) described by the tuple $(\mathcal{S}, \mathcal{A}, \mathcal{X}, \mathcal{T}, \mathcal{U}, \mathcal{R}, \rho_0, \gamma)$, where $\mathcal{S}$ is the set of states, $\mathcal{A}$ is the set of actions, and $\mathcal{X}$ is the set of observations. $\mathcal{T}:\mathcal{S}\times \mathcal{A} \rightarrow P(\mathcal{S})$ is the transition probability function where $P(\mathcal{S})$ denotes the space of probability distributions over $\mathcal{S}$, $\mathcal{U}:\mathcal{S} \rightarrow P(\mathcal{X})$ is the observation probability function, and $\mathcal{R}:\mathcal{S}\times \mathcal{A} \rightarrow \mathbb{R}$ is the reward function which maps state-action pairs to scalar rewards. Finally, $\rho_0 \in P(\mathcal{S})$ is the initial state distribution and $\gamma \in [0,1)$ the discount factor. The true environment state $s \in \mathcal{S}$ is unobserved by the agent. Given an action $a\in\mathcal{A}$, the next state is sampled such that $s'\sim\mathcal{T}(\cdot|s,a)$, an observation is generated as $x'\sim\mathcal{U}(\cdot|s')$, and a reward $\mathcal{R}(s,a)$ is computed. Note that an MDP is a special case of a POMDP where the underlying state $s$ is directly observed.

\paragraph{Reinforcement learning} Given an MDP and a stationary policy $\pi:\mathcal{S} \to P(\mathcal{A})$, the RL objective is to maximize the expected total discounted return $J(\pi)=\mathbb{E}_{\tau}[\sum_{t=0}^{\infty}\gamma^t \mathcal{R}(s_t,a_t)]$ where $\tau = (s_0,a_0,s_1,a_1,\dots)$. A stationary policy $\pi$ induces a normalized discounted state visitation distribution defined as $d_{\pi}(s) = (1-\gamma)\sum_{t=0}^{\infty}\gamma^t\mathbb{P}(s_t=s | \rho_0, \pi, \mathcal{T})$, and we define the corresponding normalized discounted state-action visitation distribution as $\rho_{\pi}(s,a)=d_{\pi}(s)\pi(a|s)$. Finally, we denote the state value function of $\pi$ as $V^{\pi}(s) = \mathbb{E}_{\tau}[\sum_{t=0}^{\infty}\gamma^t \mathcal{R}(s_t,a_t)|S_0=s]$ and the state-action value function as $Q^{\pi}(s,a) = \mathbb{E}_{\tau}[\sum_{t=0}^{\infty}\gamma^t \mathcal{R}(s_t,a_t)|S_0=s, A_0=a]$. When a function is parameterized with parameters $\bm{\theta} \in \varTheta \subset \mathbb{R}^k$ we write $\pi_{\bm{\theta}}$.

\paragraph{Generative adversarial imitation learning} Assume we have a set of expert demonstrations $\tau_E = (s_{0:T}, a_{0:T})$ generated by the expert policy $\pi_E$, a set of trajectories $\tau_{\bm{\theta}}$ generated by the policy $\pi_{\bm{\theta}}$, and a discriminator network $D_{\bm{\chi}}: \mathcal{S}\times\mathcal{A} \to [0,1]$ parameterized by $\bm{\chi}$. Generative adversarial IL \cite{ho2016generative} optimizes the min-max objective 
\begin{align}
    \min_{\bm{\theta}} \max_{\bm{\chi}} \ &\mathbb{E}_{\tau_E}[\log(D_{\bm{\chi}}(s,a))] + \mathbb{E}_{\tau_{\bm{\theta}}}[\log(1 - D_{\bm{\chi}}(s,a))]. \label{eq:IRL_disc}
\end{align}
Maximizing \eqref{eq:IRL_disc} with respect to $\bm{\chi}$ is effectively an inverse RL step where a reward function, $r_{\bm{\chi}}(s,a) = -\log(1-D_{\bm{\chi}}(s,a))$, is inferred by leveraging $\tau_E$ and $\tau_{\bm{\theta}}$. Minimizing \eqref{eq:IRL_disc} with respect to $\bm{\theta}$ is an RL step, where the agent aims to minimize its expected cost. Optimizing \eqref{eq:IRL_disc} is equivalent to minimizing $\mathbb{D}_{\text{JS}}(\rho_{\pi_{\bm{\theta}}}(s,a)||\rho_{\pi_E}(s,a))$, so we are recovering the expert state-action visitation distribution \cite{ghasemipour2020divergence}. 

\paragraph{Modeling the visual mismatch in POMDPs} Traditionally, the V-IfO problem assumes that both the expert and the agent operate within the same POMDP. Throughout this paper we relax this assumption and define two different decision processes: namely a \textit{target-POMDP} for the agent and a \textit{source-POMDP} for the expert. The target-POMDP is characterized by the tuple $(\mathcal{S}, \mathcal{A}, \mathcal{X}, \mathcal{T}, \mathcal{U}_T, \mathcal{R}, \rho_0, \gamma)$, whereas the source-POMDP is characterized by the tuple $(\mathcal{S}, \mathcal{A}, \mathcal{X}, \mathcal{T}, \mathcal{U}_S, \mathcal{R}, \rho_0, \gamma)$. The primary distinction between these POMDPs lies in their observation probability functions. Despite sharing identical state and action spaces, given the same state $s_t$, the expert's observation $x_t^S \sim \mathcal{U}_S(\cdot|s_t)$ from the source-POMDP and the agent's observation $x_t^T \sim \mathcal{U}_T(\cdot|s_t)$ from the target-POMDP may be different (i.e., we may have $x_t^S \neq x_t^T$). We refer to this as visual mismatch.

\section{Contrastive Latent Adversarial Imitation from Observations}
\label{sec:methods}

\begin{figure}
    \centering
    \includegraphics[width=\linewidth]{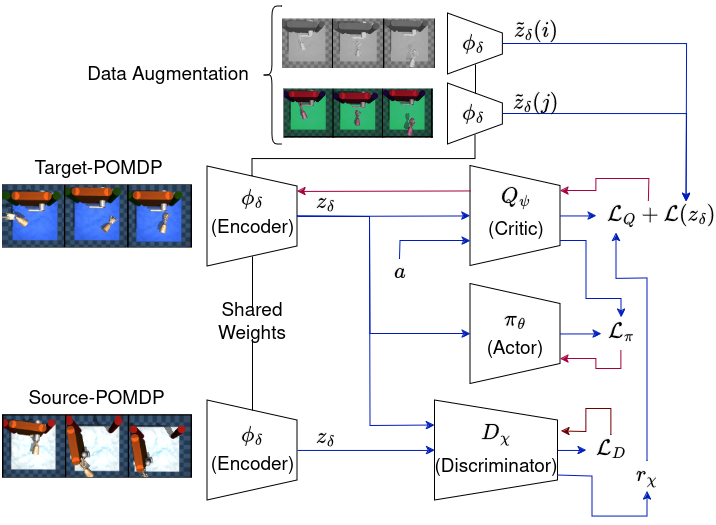}
    \caption{Summary of C-LAIfO. In the diagram, black lines indicate shared weights among networks, \textcolor{blue}{blue} arrows indicate forward pass through the networks, and \textcolor{red}{red} arrows indicate backward pass. The losses $\mathcal{L}_{D}$, $\mathcal{L}_Q$ and $\mathcal{L}(z_{\bm{\delta}})$ are respectively in \eqref{eq:AIL_BCE}, \eqref{eq:Q_regression_regularized}, and \eqref{eq:contr_loss}. $\mathcal{L}_{\pi}$ indicates the deterministic actor-critic loss~\cite{silver2014deterministic}.}
    \label{fig:C-LAIfO_scheme}
\end{figure}

Considering a target-POMDP and a source-POMDP as introduced in the previous section, we can identify two levels of information in the observation space $\mathcal{X}$: $(i)$ information related to task completion, and $(ii)$ visual distractors that do not contribute to task completion. Thus, we define $\mathcal{X}$ as $\mathcal{X} = (\bar{\mathcal{X}}, \hat{\mathcal{X}})$, where $\bar{\mathcal{X}}$ represents the \textit{goal-completion information} that is invariant between source-POMDP and target-POMDP; whereas, $\hat{\mathcal{X}}$ represents the set of \textit{visual distractors} that do not contribute to goal completion. We express the source and target observations respectively as $x_t^S = (\bar{x}_t^S, \hat{x}_t^S)$ and $x_t^T = (\bar{x}_t^T, \hat{x}_t^T)$. Our objective is to filter out the visually-distracting information $\hat{\mathcal{X}}$ from both the source and the target observations while retaining the goal-completion information $\bar{\mathcal{X}}$ to effectively solve the V-IfO problem. This objective can be achieved by attaining \textit{domain invariance in a feature space $\mathcal{Z}$}. As a result, our goal becomes to learn $\mathcal{Z}$ such that only goal-completion information is retained while the visually-distracting information is discarded (cf. supplementary material in the \href{https://arxiv.org/pdf/2407.12792}{Arxiv version} of the paper for formal analysis). 

In the following, we present the main components of our algorithm C-LAIfO, which performs imitation directly in a domain-invariant feature space $\mathcal{Z}$ (Sec.~\ref{sec:C-LAIfO_AIL}). In order to do so, we learn a domain-invariant encoder $\phi_{\bm{\delta}}$ that can successfully map $x^T_{\leq t}$ and $x^S_{\leq t}$ to $z_t$ through two main steps. First, we train the encoder $\phi_{\bm{\delta}}$ alongside the critic networks $Q_{\bm{\psi}_k}$ where $k = \{1,2\}$ (Sec.~\ref{sec:C-LAIfO_Q_phi}). This step is essential for solving the imitation problem and embedding goal-completion information within the latent space $\mathcal{Z}$. We further train $\phi_{\bm{\delta}}$ to optimize an auxiliary contrastive loss and perform randomized augmentation of the observations, taking into account the type of visual mismatch between the source and the target domains (Sec.~\ref{sec:C-LAIfO_AIL_CL}). This step is crucial to efficiently discard visually-distracting information from $\mathcal{Z}$. A schematic diagram summarizing the whole C-LAIfO pipeline is provided in Fig.~\ref{fig:C-LAIfO_scheme}.


\subsection{Adversarial imitation in latent space} 
\label{sec:C-LAIfO_AIL} 
Given a domain-invariant feature space $\mathcal{Z}$, our AIL pipeline is defined as follows. We initialize two replay buffers $\mathcal{B}_E$ and $\mathcal{B}$ to respectively store the sequences of observations generated by the expert and the agent policies, from which we infer the latent state-transitions $(z_{\bm{\delta}},z_{\bm{\delta}}')$. We write $(z_{\bm{\delta}},z_{\bm{\delta}}') \sim \mathcal{B}$ to streamline the notation. Then, given a discriminator $D_{\bm{\chi}}:\mathcal{Z}\times\mathcal{Z} \to [0,1]$, we write
\begin{align}
\begin{split}
    \max_{\bm{\chi}} \ &\mathbb{E}_{(z_{\bm{\delta}},z_{\bm{\delta}}') \sim \mathcal{B}_E}[\log(D_{\bm{\chi}}(z_{\bm{\delta}},z_{\bm{\delta}}'))] \\ &+ \mathbb{E}_{(z_{\bm{\delta}},z_{\bm{\delta}}') \sim \mathcal{B}}[\log(1 - D_{\bm{\chi}}(z_{\bm{\delta}},z_{\bm{\delta}}'))]. \label{eq:AIL_BCE}
\end{split}
\end{align} 
As mentioned, alternating \eqref{eq:AIL_BCE} with an RL step using $r_{\bm{\chi}}(z_{\bm{\delta}}, z_{\bm{\delta}}') = -\log(1-D_{\bm{\chi}}(z_{\bm{\delta}},z_{\bm{\delta}}'))$ leads to the minimization of $\mathbb{D}_{\text{JS}}\big(\rho_{\pi_{\bm{\theta}}}(z_{\bm{\delta}}, z_{\bm{\delta}}')||\rho_{\pi_{E}}(z_{\bm{\delta}}, z_{\bm{\delta}}')\big)$ \cite{goodfellow2020generative}. Therefore, we are effectively imitating the expert in the latent space $\mathcal{Z}$. Note that the presented AIL can only succeed if $\mathcal{Z}$ is domain-invariant and embeds the relevant goal-completion information necessary to solve the imitation problem. Next, we show how our algorithm C-LAIfO addresses this challenge. 

\subsection{Critic and encoder training step}
\label{sec:C-LAIfO_Q_phi}
We define the encoder as $\phi_{\bm{\delta}}:\mathcal{X}^d\to\mathcal{Z}$, a function mapping sequences of $d \in \mathbb{N}$ observations to the latent space $\mathcal{Z}$. Specifically, we write $z_{\bm{\delta}} = \phi_{\bm{\delta}}(x_{t^-:t})$ and $z_{\bm{\delta}}' = \phi_{\bm{\delta}}(x_{t^-+1:t+1})$ where $t-t^-+1 = d$. When a data augmentation function $\text{aug}(\cdot)$ is applied to the sequence of observations, we write $\tilde z_{\bm{\delta}} = \phi_{\bm{\delta}}(\text{aug}(x_{t^-:t}))$ and $\tilde z_{\bm{\delta}}' = \phi_{\bm{\delta}}(\text{aug}(x_{t^-+1:t+1}))$. We train $\phi_{\bm{\delta}}$ to optimize
\begin{align}
\begin{split}
    \min_{\bm{\psi}_k, \bm{\delta}}& \ \mathbb{E}_{(\tilde z_{\bm{\delta}}, a_t, \tilde z_{\bm{\delta}}')\sim\mathcal{B}}[\big(Q_{\bm{\psi}_k}(\tilde z_{\bm{\delta}}, a_t) - \text{sg}(y)\big)^2] \\ &\ \ \ + \mathbb{E}_{\tilde z_{\bm{\delta}} \sim \mathcal{B}}[\mathcal{L}(\tilde z_{\bm{\delta}})] \label{eq:Q_regression_regularized}
    \end{split}\\
    \text{s.t.}& \ y = r_{\bm{\chi}}(z_{\bm{\delta}}, z_{\bm{\delta}}') + \gamma \min_{k={1,2}} Q_{\bar{\bm{\psi}}_k}(\tilde z_{\bm{\delta}}', a'). \label{eq:Bellman_operator}
\end{align}

The steps in \eqref{eq:Q_regression_regularized}--\eqref{eq:Bellman_operator} follow the deep $Q$-network optimization pipeline \cite{mnih2013playing, mnih2015human}, where we add a contrastive auxiliary loss $\mathbb{E}_{\tilde z_{\bm{\delta}} \sim \mathcal{B}}[\mathcal{L}(\tilde z_{\bm{\delta}})]$ in \eqref{eq:Q_regression_regularized} defined on the encoder $\phi_{\bm{\delta}}$. In \eqref{eq:Bellman_operator}, the reward function $r_{\bm{\chi}}(z_{\bm{\delta}}, z_{\bm{\delta}}')$ is computed through the AIL step in \eqref{eq:AIL_BCE}. Note that we do not perform data augmentation when computing $r_{\bm{\chi}}$ since, during reward inference, we are deploying and not training the encoder $\phi_{\bm{\delta}}$. In practice, adding data augmentation to AIL in \eqref{eq:AIL_BCE} decreases the performance. Refer to the \href{https://arxiv.org/pdf/2407.12792}{Arxiv version} for empirical evidence on this claim. In \eqref{eq:Q_regression_regularized}, the encoder $\phi_{\bm{\delta}}$ is trained together with the critic networks $Q_{\bm{\psi}_k}$ ($k=\{1,2\}$) in order to regress $y$ in \eqref{eq:Bellman_operator}, where $\text{sg}(\cdot)$ stands for stop gradient of the encoder parameters $\bm{\delta}$. The value of $y$ in \eqref{eq:Bellman_operator} is computed by summing $r_{\bm{\chi}}$ with the discounted target critic network at time $t+1$. In \eqref{eq:Bellman_operator}, $a' = \pi_{\bm{\theta}}(\tilde z_{\bm{\delta}}') + \epsilon$ where $\epsilon \sim \text{clip}(\mathcal{N}(0,\sigma^2), -c, c)$ is a clipped exploration noise with $c$ the clipping parameter and $\mathcal{N}(0,\sigma^2)$ a univariate normal distribution with zero mean and $\sigma$ standard deviation. $\bar{\bm{\psi}}_1$ and $\bar{\bm{\psi}}_2$ are the slow moving weights for the target critic networks. $\mathcal{B}$ is a replay buffer initialized to store interactions $(x^T_t, a_t, x^T_{t+1})$ of the agent with the target environment. Note that the latent state-transitions $(\tilde z_{\bm{\delta}}, \tilde z_{\bm{\delta}}')$ are inferred from sequences of observations using $\phi_{\bm{\delta}}$ and, as above, we write $(\tilde z_{\bm{\delta}}, \tilde z_{\bm{\delta}}') \sim \mathcal{B}$ to streamline the notation. Refer to the \href{https://arxiv.org/pdf/2407.12792}{Arxiv version} for the complete pseudo-code. 

By solving the optimization problem in \eqref{eq:Q_regression_regularized}--\eqref{eq:Bellman_operator}, our primary goal is to train both the encoder network $\phi_{\bm{\delta}}$ and the critic networks $Q_{\bm{\psi}_k}$ to solve the RL problem with reward $r_{\bm{\chi}}$. In other words, this step focuses on retaining the goal-completion information within the latent space $\mathcal{Z}$ such that critic networks $Q_{\bm{\psi}_k}$ are successfully learned. We show in our ablation study that backpropagating the gradient from $Q_{\bm{\psi}_k}$ to $\phi_{\bm{\delta}}$ is an important step for achieving this goal and solving the imitation problem. Similarly, the types of augmentations performed on the sequences of observations and the choice of auxiliary loss play a crucial role in discarding the visually-distracting information from $\mathcal{Z}$ and dealing with the visual mismatch. 

\subsection{Contrastive loss}
\label{sec:C-LAIfO_AIL_CL}
In the following, we introduce the data augmentation techniques and the auxiliary loss $\mathcal{L}$ in \eqref{eq:Q_regression_regularized} for C-LAIfO. We opt for a contrastive method as this leads to good empirical results and good computational efficiency. Contrastive learning constructs low-dimensional representations of high-dimensional data by maximizing agreement between augmented views of the same data example via a contrastive loss in the latent space $\mathcal{Z}$. In our specific case, we define equivalent data as sequences of observations with the same goal-completion information. The data augmentation is randomized by considering a set of pre-determined functions. We will show in our experiments (Section~\ref{sec:experiments}) that both the choice of contrastive loss and the set of augmentation functions play an important role in filtering out visually-distracting information. 

First, a stochastic data augmentation module transforms any given sequence of observations $x_{t^-:t}$ into two views, denoted $\text{aug}(x_{t^-:t})_i$ and $\text{aug}(x_{t^-:t})_j$, which are denoted as the positive pairs. Note that two positive pairs must contain the same goal-completion information. Next, the encoder $\phi_{\bm{\delta}}:\mathcal{X}^d\to\mathcal{Z}$ extracts representation vectors from augmented sequences of observations. We write $\tilde z_{\bm{\delta}}(i) = \phi_{\bm{\delta}}(\text{aug}(x_{t^-:t})_i)$ and $\tilde z_{\bm{\delta}}(j) = \phi_{\bm{\delta}}(\text{aug}(x_{t^-:t})_j)$. Finally, we apply the contrastive loss function
\begin{equation}
    \mathcal{L}(z_{\bm{\delta}}) = -\log \frac{\exp(\text{sim}(\tilde z_{\bm{\delta}}(i), \tilde z_{\bm{\delta}}(j))/\eta)}{\sum_{k=1}^{2N} \mathds{1}_{[k\neq i]} \exp(\text{sim}(\tilde z_{\bm{\delta}}(i), \tilde z_{\bm{\delta}}(k))/\eta)},
    \label{eq:contr_loss}
\end{equation}
where $\mathds{1}_{[k\neq i]} \in \{0,1\}$ is an indicator function equal to 1 if $k\neq i$, $\eta$ denotes a temperature parameter, and $\text{sim}(\bm{u}, \bm{v}) = \bm{u}^{\intercal}\bm{v}/||\bm{u}||||\bm{v}||$ denotes the cosine similarity. We sample a batch of $N$ sequences of observations from the buffer $\mathcal{B}$ and define pairs of augmented sequences derived from the batch, resulting in $2N$ data points. The negative data points are not sampled explicitly. Instead, given a positive pair, we treat the other $2(N-1)$ augmented data points within the batch as negatives. Note that the loss in \eqref{eq:contr_loss} is called the normalized temperature-scaled cross entropy loss or the Information Noise-Contrastive Estimation (InfoNCE) loss \cite{oord2018representation} and represents an upper bound of the negative mutual information between positive pairs. Therefore, by minimizing \eqref{eq:contr_loss} as in \eqref{eq:Q_regression_regularized} we are maximizing the mutual information between positive pairs in the latent space $\mathcal{Z}$.   

\section{Experiments}
\label{sec:experiments}
In this section, we begin with an ablation study to justify the design choices of our algorithm (Sec.~\ref{sec:experiments_ablation_studies}). Next, we demonstrate how C-LAIfO effectively handles various types of visual mismatches in the V-IfO setting (Sec.~\ref{sec:experiments_V-IfO}). Finally, we showcase how C-LAIfO facilitates learning in challenging robotic manipulation tasks with sparse rewards and realistic visual inputs (Sec.~\ref{sec:experiments_RL+IL}). In all the experiments, we use DDPG~\cite{lillicrap2015continuous} to train experts in a fully observable setting and collect $100$ episodes of expert data. The learned policies are evaluated based on average return over $10$ episodes. We report the mean and standard deviation of the final return over $6$ seeds and \textbf{highlight} the best performance. 

\subsection{Ablation study}
\label{sec:experiments_ablation_studies}

\begin{table}
\centering
\footnotesize
\caption{Summary of the ablation experiments. All algorithms are trained for $10^6$ steps. The experiments \textit{C-LAIfO w/o $Q$ backprop}, \textit{C-LAIfO full aug}, and \textit{C-LAIfO w/o aug} are only conducted in the easier setting due to their low performance.}
\label{table_ablation}
    \begin{tabular}{c | c c}
        \toprule
        \multicolumn{3}{c}{\makecell{Source Env: \raisebox{-.5\height}{\includegraphics[width=0.12\linewidth]{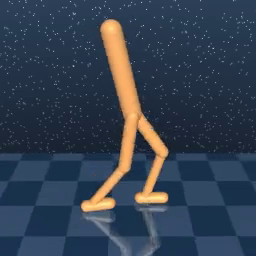}}, Performance = $950$}} \\
        \cmidrule(lr){1-3}
        Target Env & \raisebox{-.5\height}{\includegraphics[width=0.12\linewidth]{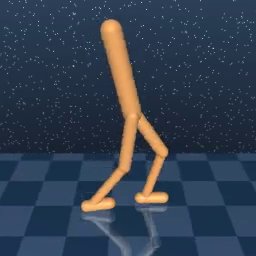}} &  \raisebox{-.5\height}{\includegraphics[width=0.12\linewidth]{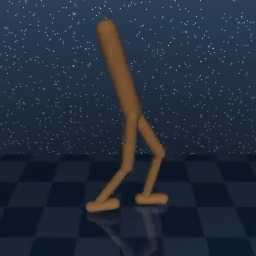}} \\
        \cmidrule(lr){1-3}
        \makecell{C-LAIfO} & \bm{$808 \pm 269$} & \bm{$768 \pm 231$} \\
        \makecell{BYOL-LAIfO} & $707 \pm 337$ & $142 \pm 124$ \\
        \makecell{C-LAIfO w/o $Q$ backprop} & $48.7 \pm 8.7$ & - \\
        \makecell{C-LAIfO full aug} & $96.8 \pm 53.4$ & - \\
        \makecell{C-LAIfO w/o aug} & $113 \pm 25.3$ & - \\     
        \bottomrule
    \end{tabular}
\end{table}

In this section, we perform the following ablations:
\begin{enumerate}
    \item \textbf{Contrastive loss function}: We demonstrate the importance of the contrastive loss function type by comparing the InfoNCE loss in \eqref{eq:contr_loss} with BYOL in \cite{grill2020bootstrap}.

    \item \textbf{Gradient backpropagation}: We highlight the necessity of backpropagating the gradient from $Q_{\bm{\psi}_k}$ to $\phi_{\bm{\delta}}$ in \eqref{eq:Q_regression_regularized} for solving the imitation problem and embedding the goal-completion information in the latent space $\mathcal{Z}$.

    \item \textbf{Data augmentation}: We emphasize the importance of selecting the appropriate augmentation for a given mismatch, showing that a mismatch-informed augmentation outperforms general augmentations or no augmentation. 
\end{enumerate}

These results are summarized in Table~\ref{table_ablation}, which includes results for C-LAIfO and its various configurations. All the learning curves are provided in the \href{https://arxiv.org/pdf/2407.12792}{Arxiv version} of the paper. In Table~\ref{table_ablation}, \textit{C-LAIfO} is implemented as described in Section~\ref{sec:methods}, with the data augmentation function $\text{aug}(\cdot)$ defined as a brightness transformation. In \textit{BYOL-LAIfO}, we retain the design identical to C-LAIfO except for replacing the InfoNCE loss in \eqref{eq:contr_loss} with BYOL \cite{grill2020bootstrap}. In \textit{C-LAIfO w/o $Q$ backprop}, we disable gradient backpropagation from $Q_{\bm{\psi}k}$ to $\phi_{\bm{\delta}}$ in \eqref{eq:Q_regression_regularized}. Lastly, in \textit{C-LAIfO full aug} and \textit{C-LAIfO w/o aug}, we respectively modify the data augmentation function $\text{aug}(\cdot)$ to include full augmentation (brightness, color, and geometric transformations, as detailed in the \href{https://arxiv.org/pdf/2407.12792}{Arxiv version} of the paper) and no augmentation. Notably, without backpropagating the gradient from $Q_{\bm{\psi}_k}$ to $\phi_{\bm{\delta}}$, C-LAIfO fails to solve the imitation task even in the simplest mismatch scenario. This result demonstrates the importance of this step for embedding goal-completion information in $\mathcal{Z}$. Similar considerations can be done for the design of $\text{aug}(\cdot)$ where C-LAIfO struggles to efficiently solve the task both when the augmentation is too generic and when it is absent. These results highlight the critical role of properly defining $\text{aug}(\cdot)$ for efficient visual generalization in $\mathcal{Z}$. Finally, the superior performance of the InfoNCE loss in \eqref{eq:contr_loss} compared to BYOL is evident in handling the hardest mismatch in Table~\ref{table_ablation}.

\subsection{Visual Imitation from Observations with mismatch}
\label{sec:experiments_V-IfO}
\begin{figure}
    \centering
    \begin{subfigure}[t]{0.22\linewidth}
        \centering
        \includegraphics[width=\linewidth]{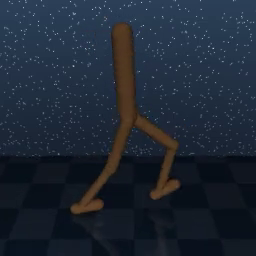}
        \caption{}
        \label{fig:walker_dark}
    \end{subfigure}
    \begin{subfigure}[t]{0.22\linewidth}
        \centering
        \includegraphics[width=\linewidth]{Figures/walker_walk_delta=0.20.png}
        \caption{}
        \label{fig:walker_standard}
    \end{subfigure}
    \begin{subfigure}[t]{0.22\linewidth}
        \centering
        \includegraphics[width=\linewidth]{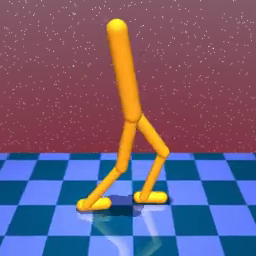}
        \caption{}
        \label{fig:walker_full}
    \end{subfigure}
    \begin{subfigure}[t]{0.22\linewidth}
        \centering
        \includegraphics[width=\linewidth]{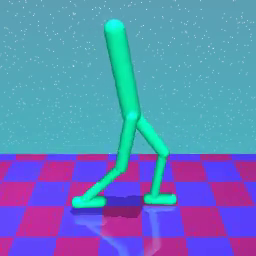}
        \caption{}
        \label{fig:walker_unseen}
    \end{subfigure}
    \caption{Different environments used for the experiments in Table~\ref{table_visual_experiments} and the PCA in Fig.~\ref{fig:walker_PCA_light} and \ref{fig:walker_PCA_full}.}
    \label{fig:color_walker_walk_mismatch}
\end{figure}

\begin{table}[h!]
\centering
\footnotesize
\caption{Summary of the experiments for the mismatches in Fig.~\ref{fig:color_walker_walk_mismatch}. The Light experiment consists in \eqref{fig:walker_standard} as source-POMDP and \eqref{fig:walker_dark} as target-POMDP. The Full experiments have \eqref{fig:walker_standard} as target-POMDP and \eqref{fig:walker_full} as source-POMDP. We train all the algorithms in the Light mismatch for $10^6$ steps and in the Full mismatch for $2\times10^6$ steps.}
\label{table_visual_experiments}
    \begin{tabular}{c | c c}
        \toprule
        & Light & Full \\
        \cmidrule(lr){1-3}
        \makecell{C-LAIfO} & \bm{$895 \pm 36.6$} & \bm{$509 \pm 235$} \\
        LAIfO~\cite{giammarino2023adversarial} w/ data aug & $64.6 \pm 62.9$ & $206 \pm 210$ \\
        DisentanGAIL~\cite{cetin2020domain} & $28.1 \pm 8.7$ & $30.6 \pm 13.2$ \\
        PatchAIL~\cite{liu2022visual} w/ data aug & $18.4 \pm 5.6$ & $122 \pm 66.6$ \\
        \bottomrule
    \end{tabular}
\end{table}

\begin{figure}[h!]
    \centering
    \includegraphics[width=0.8\linewidth]{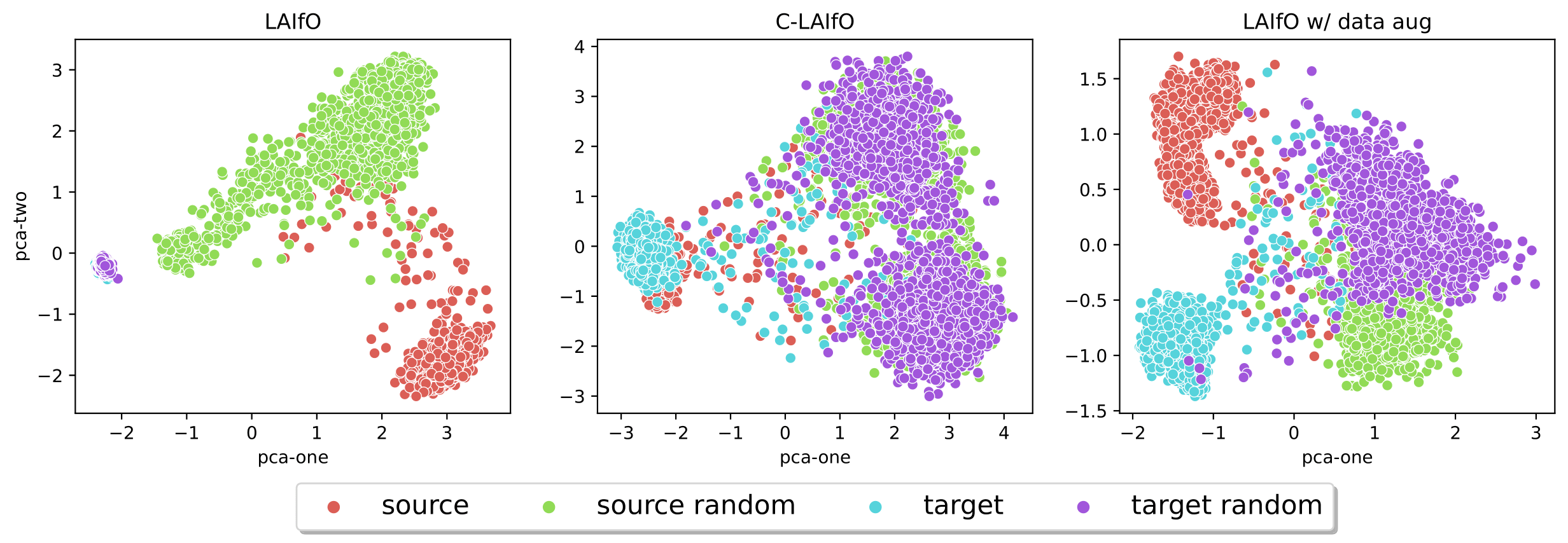}
    \caption{PCA results for the Light experiment in Table~\ref{table_visual_experiments}.}
    \label{fig:walker_PCA_light}
\end{figure}

\begin{figure}[h!]
    \centering
    \includegraphics[width=0.6\linewidth]{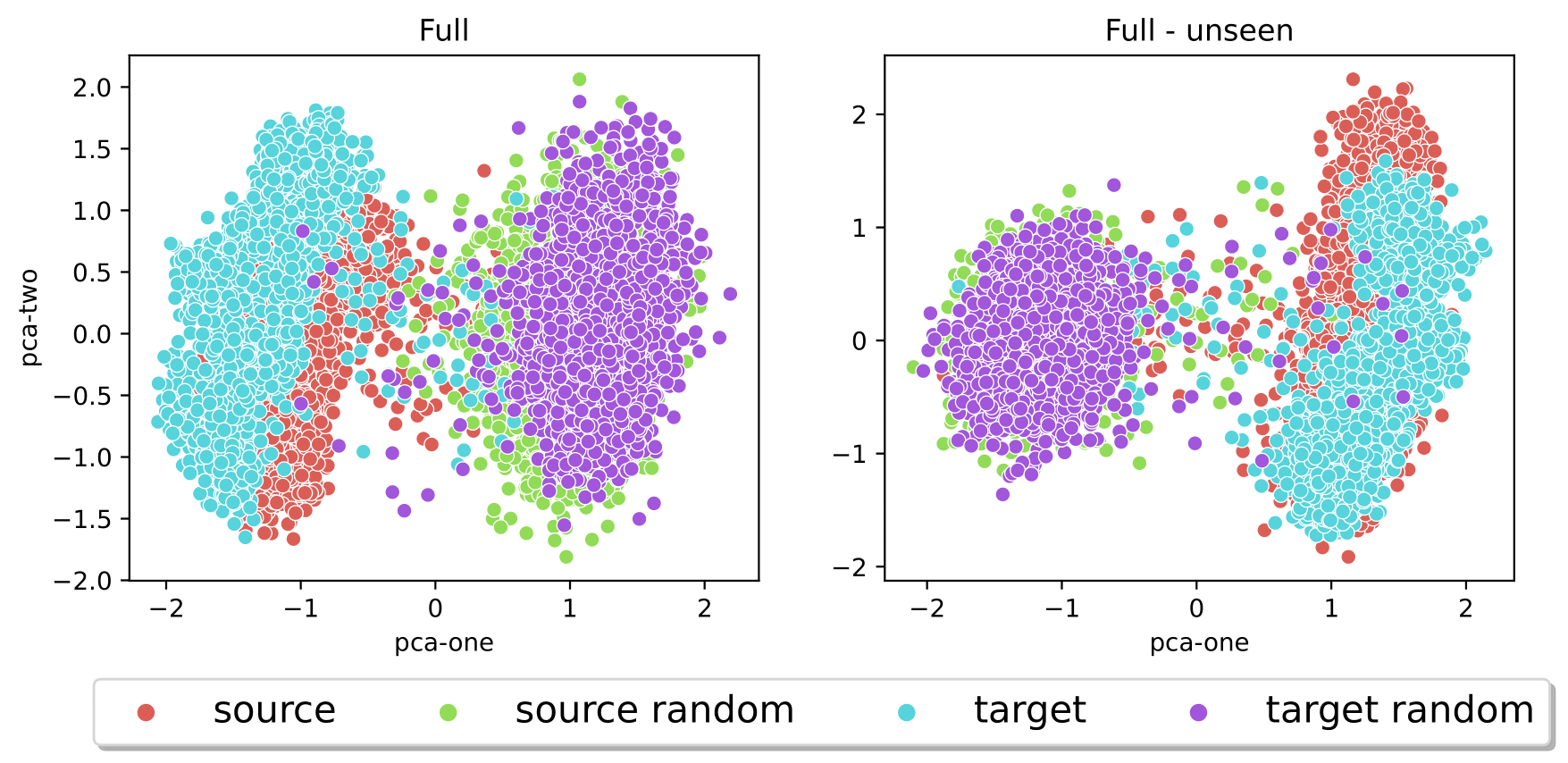}
    \caption{PCA results on C-LAIfO for the Full experiment in Table~\ref{table_visual_experiments} and the unseen environment in Fig.~\ref{fig:walker_unseen}.}
    \label{fig:walker_PCA_full}
\end{figure}

\begin{figure*}[ht!]
    \centering
    \begin{subfigure}[t]{0.1\linewidth}
        \centering
        \includegraphics[width=\linewidth]{Figures/Door.png}
        \caption{}
        \label{fig:door}
    \end{subfigure}
    \begin{subfigure}[t]{0.1\linewidth}
        \centering
        \includegraphics[width=\linewidth]{Figures/door_mismatch.png}
        \caption{}
        \label{fig:door-light}
    \end{subfigure}
    \begin{subfigure}[t]{0.1\linewidth}
        \centering
        \includegraphics[width=\linewidth]{Figures/Door_color.png}
        \caption{}
        \label{fig:door-color}
    \end{subfigure}
    \begin{subfigure}[t]{0.1\linewidth}
        \centering
        \includegraphics[width=\linewidth]{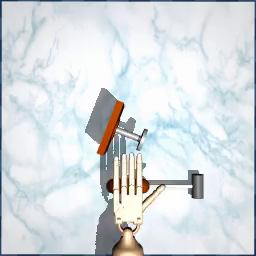}
        \caption{}
        \label{fig:hammer}
    \end{subfigure}
    \begin{subfigure}[t]{0.1\linewidth}
        \centering
        \includegraphics[width=\linewidth]{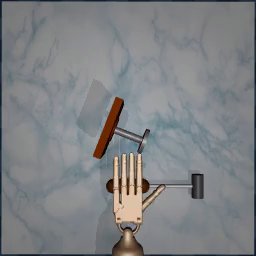}
        \caption{}
        \label{fig:hammer-light}
    \end{subfigure}
    \begin{subfigure}[t]{0.1\linewidth}
        \centering
        \includegraphics[width=\linewidth]{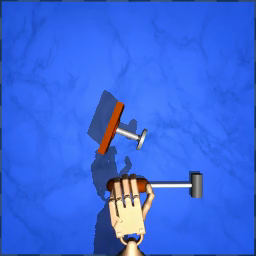}
        \caption{}
        \label{fig:hammer-color}
    \end{subfigure}
    \begin{subfigure}[t]{0.1\linewidth}
        \centering
        \includegraphics[width=\linewidth]{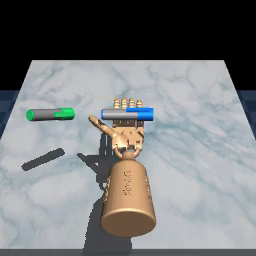}
        \caption{}
        \label{fig:pen}
    \end{subfigure}
    \begin{subfigure}[t]{0.1\linewidth}
        \centering
        \includegraphics[width=\linewidth]{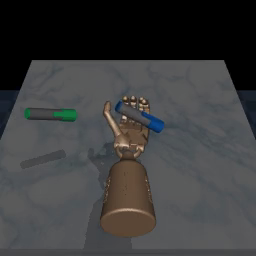}
        \caption{}
        \label{fig:pen-light}
    \end{subfigure}
    \begin{subfigure}[t]{0.1\linewidth}
        \centering
        \includegraphics[width=\linewidth]{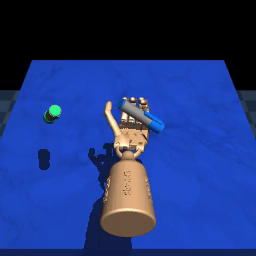}
        \caption{}
        \label{fig:pen-color}
    \end{subfigure}
    \caption{Adroit environments used for the experiments in Table~\ref{table_adroit}.}
    \label{fig:adroit}
\end{figure*}

\begin{table*}[ht!]
\centering
\caption{Summary of the experiments in Fig.~\ref{fig:adroit}. The Door-Light and Door-Color experiments consider \eqref{fig:door} as the source-POMDP and \eqref{fig:door-light} and \eqref{fig:door-color} as the respective target-POMDPs. Similarly, the Hammer-Light and Hammer-Color experiments consider \eqref{fig:hammer} as source-POMDP and \eqref{fig:hammer-light} and \eqref{fig:hammer-color} as the respective target-POMDPs, while the Pen-Light and Pen-Color experiments consider \eqref{fig:pen} as source-POMDP and \eqref{fig:pen-light} and \eqref{fig:pen-color} as the respective target-POMDPs. We use the VRL3 pipeline in \cite{wang2022vrl3} to train expert policies and collect $100$ episodes of expert data. For these experiments, all the algorithms are trained for $4 \times 10^6$ steps and we report mean and standard deviation over 10 random seeds.}
\label{table_adroit}
    \begin{tabular}{c | c c | c c | c c}
        \toprule
        & \multicolumn{2}{c|}{Door} & \multicolumn{2}{c|}{Hammer} & \multicolumn{2}{c}{Pen} \\
        & Light & Color & Light & Color & Light & Color \\
        \cmidrule(lr){1-7}
        Expert & \multicolumn{2}{c|}{$170$} & \multicolumn{2}{c|}{$184$} & \multicolumn{2}{c}{$73$} \\
        \cmidrule(lr){1-7}
        RL+LAIfO~\cite{giammarino2023adversarial} & $106 \pm 75$ & $96 \pm 80$ & $181 \pm 4.0$ & $103 \pm 86$ & $59 \pm 12$ & $59 \pm 4.8$ \\
        \cmidrule(lr){1-7}
        RL+C-LAIfO & \bm{$165 \pm 5.8$} & \bm{$160 \pm 12$} & \bm{$183 \pm 1.4$} & \bm{$178 \pm 11$} & \bm{$64 \pm 5.6$} & \bm{$62 \pm 6.9$}  \\
        \bottomrule
    \end{tabular}
\end{table*}

In this section, we test C-LAIfO in the V-IfO setting with different types of mismatches (cf. Fig.~\ref{fig:color_walker_walk_mismatch}) and compare it with three baselines: LAIfO~\cite{giammarino2023adversarial} and PatchAIL~\cite{liu2022visual}, both equipped with the same $\text{aug}(\cdot)$ used for C-LAIfO, and DisentanGAIL~\cite{cetin2020domain}. The results, summarized in Table~\ref{table_visual_experiments}, demonstrate that C-LAIfO successfully addresses the V-IfO with mismatch problem, achieving superior performance compared to all the baselines across the proposed mismatches. All the learning curves are provided in the \href{https://arxiv.org/pdf/2407.12792}{Arxiv version} of the paper. For the \textit{Light} experiment, $\text{aug}(\cdot)$ is defined as a brightness transformation; while for the others it is defined as a color transformation. Details on $\text{aug}(\cdot)$ are provided in the \href{https://arxiv.org/pdf/2407.12792}{Arxiv version} of the paper. Furthermore, to assess whether C-LAIfO achieves \textit{domain invariance in the feature space $\mathcal{Z}$}, we perform PCA on the latent space $\mathcal{Z}$ learned by different algorithms during training. In Fig.~\ref{fig:walker_PCA_light} we compare the latent space $\mathcal{Z}$ learned by LAIfO, C-LAIfO, and LAIfO with data augmentation in the Light setting from Table~\ref{table_visual_experiments}. Specifically, we define \textit{source} and \textit{target} as the observations generated by an optimal policy in the source and target POMDPs, respectively. Similarly, \textit{source random} and \textit{target random} are observations generated by random policies. These observations are processed with the encoder $\phi_{\bm{\delta}}:\mathcal{X}^d \to \mathcal{Z}$ trained using the respective algorithms. We perform PCA on this set of latent variables $z_{1:T}$ and plot the first two principal components. The results show that C-LAIfO is the only algorithm capable of filtering out visual distractors and clustering together data points with the same goal-completion information. In Fig.~\ref{fig:walker_PCA_full}, we focus on C-LAIfO and test for generalization to the unseen environment in Fig.~\ref{fig:walker_unseen}. We train $\phi_{\bm{\delta}}$ on the Full setting from Table~\ref{table_visual_experiments} and perform PCA as described. The results in the unseen setting match those in the Full experiment, indicating that $\phi_{\bm{\delta}}$ trained on \eqref{fig:walker_full} can successfully generalize to \eqref{fig:walker_unseen}. Refer to the \href{https://arxiv.org/pdf/2407.12792}{Arxiv version} of the paper also for the t-SNE visualization and additional details. 

\subsection{C-LAIfO for dexterous manipulation} 
\label{sec:experiments_RL+IL}

In the following section, we evaluate our algorithm on a series of challenging robotic manipulation tasks from the Adroit platform for dynamic dexterous manipulation~\cite{Kumar2016thesis}. These experiments demonstrate how the reward $r_{\bm{\chi}}$, learned by C-LAIfO from expert videos, can be effectively combined with a sparse reward $\mathcal{R}$, collected by the agent through interaction with the environment, to enhance learning efficiency. The RL problem, therefore, aims to maximize the total reward $\mathcal{R}_{\text{tot}} = \mathcal{R}(s_t,a_t) + r_{\bm{\chi}}(z_t, z_{t+1})$, where $r_{\bm{\chi}}$ is learned through the AIL step in \eqref{eq:AIL_BCE}. This approach is particularly relevant for robotic tasks, where sparse rewards are often the most feasible option in real-world settings. However, relying solely on sparse rewards can make learning highly challenging and inefficient \cite{giammarino2023reinforcement}. In this context, leveraging expert videos can significantly enhance efficiency. We compare C-LAIfO with the standard LAIfO algorithm \cite{giammarino2023adversarial}, which does not explicitly address the visual mismatch between source and target POMDPs. Both C-LAIfO and LAIfO utilize an encoder to process pixel observations and extract embeddings in $\mathcal{Z}$, which are then concatenated with robot sensory observations. Notably, the expert's sensory observations are not used in the imitation process, as we only assume access to expert videos. Our approach, denoted as RL+C-LAIfO (or RL+LAIfO), seeks to maximize $\mathcal{R}_{\text{tot}}$, rather than just $r_{\bm{\chi}}$, as in the standard imitation learning problem. The results, summarized in Table~\ref{table_adroit}, show that C-LAIfO more effectively leverages expert videos with visual mismatches to facilitate learning when compared to LAIfO. This demonstrates the potential of our approach to enable learning in challenging robotic tasks by utilizing a minimal form of supervision, relying solely on expert videos. All learning curves are provided in the \href{https://arxiv.org/pdf/2407.12792}{Arxiv version} of the paper.

\section{Conclusion}
\label{sec:conclusion}

In this work, we analyze the V-IfO problem with visual mismatches and propose a novel algorithm named C-LAIfO as an effective solution. Through comprehensive ablation studies, we provide insights into our design and demonstrate the superior performance of our approach compared to a range of baselines in imitation from videos under various mismatch scenarios. Furthermore, we illustrate how C-LAIfO effectively utilizes expert videos with visual mismatches to facilitate learning in challenging hand manipulation tasks characterized by sparse rewards and realistic visual inputs. 

A main limitation of the current approach is given by the reliance of C-LAIfO on a well-designed, possibly mismatch-informed, data augmentation function. As illustrated in our ablations in Section~\ref{sec:experiments}, general augmentation can lead to poor performance or can remarkably reduce the algorithmic sample efficiency. Furthermore, it can be challenging to design effective augmentations for certain types of mismatches. To address this problem, exploring generative models for automatic data augmentation represents an interesting research direction. Generative models could produce diverse, mismatch-informed augmentations, potentially overcoming the limitations of manually designed strategies. Alternatively, investigating different auxiliary losses that are less reliant on augmentation techniques represents another interesting direction. Finally, future work will be devoted to go beyond simulated environments and test our algorithms on hardware in real-world scenarios.  




\bibliography{Bibliography}

\clearpage
\section*{Supplementary material}

\subsection{Analysis}
\label{app_sec:theo_analysis} 

In the following, we show how \textit{domain invariance in the feature space $\mathcal{Z}$} leads to the same V-IfO guarantees as in \cite{giammarino2023adversarial}, even in the presence of visual mismatches.

\begin{proposition}
    \label{lemma_2}
Consider source and target POMDPs respectively defined by the tuples $(\mathcal{S}, \mathcal{A}, \mathcal{X}, \mathcal{T}, \mathcal{U}_T, \mathcal{R}, \rho_0, \gamma)$ and $(\mathcal{S}, \mathcal{A}, \mathcal{X}, \mathcal{T}, \mathcal{U}_S, \mathcal{R}, \rho_0, \gamma)$. Let $\mathcal{X} = (\bar{\mathcal{X}}, \hat{\mathcal{X}})$, where $\bar{\mathcal{X}}$ is an observations set invariant between source and target POMDP, and $\hat{\mathcal{X}}$ is a set of visual distractors. We write $x_t = (\bar{x}_t, \hat{x}_t)$. 
Assume $z_t = \phi(x_{\leq t}) = \phi(\bar{x}_{\leq t})$ such that $\mathbb{P}(s_t|z_t, a_t) = \mathbb{P}(s_t|z_t) = \mathbb{P}(s_t|x_{\leq t}, a_{<t}) = \mathbb{P}(s_t|\bar{x}_{\leq t}, a_{<t})$. Then, the filtering posterior distributions $\mathbb{P}(s_t|z_t)$ and $\mathbb{P}(s_{t+1},s_t|z_{t+1},z_t)$ do not depend on the policy $\pi$ and are invariant between source and target POMDPs. 

\begin{proof}
    Considering the definition of the latent variable $z_t$, which only depends on $\bar{x}_{\leq t}$ and not the visually-distracting information $\hat{x}_{\leq t}$, we can write 
    \begin{align}
        \mathbb{P}(s_t|z_t) = \mathbb{P}(s_t|x_t, a_{t-1}, z_{t-1}) = \mathbb{P}(s_t|\bar{x}_t, a_{t-1}, z_{t-1}). \nonumber
    \end{align}
    Then, by leveraging Bayes rule we have that
        \begin{align}
            \mathbb{P}&(s_t|z_t) = \mathbb{P}(s_t|\bar{x}_t, a_{t-1}, z_{t-1}) \nonumber \\ 
            &= \frac{\mathbb{P}(\bar{x}_t|s_t, a_{t-1}, z_{t-1})\mathbb{P}(s_t| a_{t-1}, z_{t-1})}{\mathbb{P}(\bar{x}_t|a_{t-1}, z_{t-1})} \nonumber \\
            &= \frac{\mathbb{P}(\bar{x}_t|s_t)\int_{\mathcal{S}}\mathcal{T}(s_t|s_{t-1},a_{t-1}) \mathbb{P}(s_{t-1}|z_{t-1}) ds_{t-1}}{\int_{\mathcal{S}}\int_{\mathcal{S}}\mathbb{P}(\bar{x}_t|s_t)\mathcal{T}(s_t|s_{t-1},a_{t-1}) \mathbb{P}(s_{t-1}|z_{t-1}) ds_t ds_{t-1}}, \nonumber
        \end{align}
        where the denominator can be seen as a normalizing factor. Note that $\mathbb{P}(\bar{x}_t|s_t)$ is the same for both the source and target POMDPs by the definition of $\bar{\mathcal{X}}$ above. Therefore, $\mathbb{P}(s_t|z_t)$ has no dependence on the policy $\pi$ and is invariant between source and target POMDP. 
        
        Similarly, for $\mathbb{P}(s_{t+1},s_t|z_t,z_{t+1})$ we have that
        \begin{align}
            \mathbb{P}&(s_{t+1},s_t|z_t,z_{t+1}) = \mathbb{P}(s_t, s_{t+1}|\bar{x}_{t+1}, a_t, z_t) \nonumber \\
            &= \frac{\mathbb{P}(\bar{x}_{t+1}|s_t, s_{t+1}, a_t, z_t)\mathbb{P}(s_t, s_{t+1}|a_t, z_t)}{\mathbb{P}(\bar{x}_{t+1}|a_t, z_t)} \nonumber \\
            &= \frac{\mathbb{P}(\bar{x}_{t+1}|s_{t+1})\mathcal{T}(s_{t+1}|s_{t},a_{t}) \mathbb{P}(s_{t}|z_{t})}{\int_{\mathcal{S}}\int_{\mathcal{S}}\mathbb{P}(\bar{x}_{t+1}|s_{t+1})\mathcal{T}(s_{t+1}|s_{t},a_{t}) \mathbb{P}(s_{t}|z_{t}) ds_{t+1} ds_{t}}. \nonumber
        \end{align}
        Because $\mathbb{P}(s_t|z_t)$ does not depend on $\pi$ and is the same for both source and target POMDP, the result also holds for $\mathbb{P}(s_{t+1},s_t|z_t,z_{t+1})$.
\end{proof}
\end{proposition}

\begin{proposition}
    \label{theorem_1}
Given $\mathcal{R}: \mathcal{S} \times \mathcal{A} \to \mathbb{R}$, for the scenarios described in Proposition~\ref{lemma_2} the following inequality holds:
\begin{align*}
    \begin{split}
        \big|J(\pi_{E}) - J(\pi_{\bm{\theta}})\big| \leq& \frac{2 R_{\max}}{1-\gamma}\mathbb{D}_{\text{\normalfont{TV}}}\big(\rho_{\pi_{\bm{\theta}}}(z,z'),\rho_{\pi_{E}}(z, z')\big) + C,
    \end{split}
\end{align*}
where $R_{\max} = \max_{(s,a) \in \mathcal{S} \times \mathcal{A}}|\mathcal{R}(s,a)|$ and 
\begin{equation}
    C = \frac{2 R_{\max}}{1-\gamma}\mathbb{E}_{\rho_{\pi_{\bm{\theta}}}(z, z')}\big[\mathbb{D}_{\text{\normalfont{TV}}}\big(\mathbb{P}_{\pi_{\bm{\theta}}}(a|z, z'),\mathbb{P}_{\pi_{E}}(a|z, z')\big)\big].
    \label{app:theo_1_c}
\end{equation}

If $\mathcal{R}: \mathcal{S} \times \mathcal{S} \to \mathbb{R}$, then we have that 
\begin{align*}
    \begin{split}
        \big|J(\pi_{E}) - J(\pi_{\bm{\theta}})\big| \leq& \frac{2 R_{\max}}{1-\gamma}\mathbb{D}_{\text{\normalfont{TV}}}\big(\rho_{\pi_{\bm{\theta}}}(z,z'),\rho_{\pi_{E}}(z, z')\big),
    \end{split}
\end{align*}
where $R_{\max} = \max_{(s,s') \in \mathcal{S} \times \mathcal{S}}|\mathcal{R}(s,s')|$.

\begin{proof}
Because Proposition~\ref{lemma_2} holds, we can directly follow the proofs of Theorem~1 and Theorem~2 in \cite{giammarino2023adversarial} for the setting of no visual mismatch.
\end{proof}
\end{proposition}

\subsection{Pseudo-code and hyperparameters}
\label{app:sec_hyperparams}

\begin{algorithm}
\small
\label{alg:C-LAIfO}
\caption{C-LAIfO}
\textbf{Inputs}: \\
Expert observations: $(x^S_n)_{0:N} \in \mathcal{B}_{E}$. \\
$\pi_{\bm{\theta}}$, $D_{\bm{\chi}}$, $Q_{\bm{\psi}_1}$, $Q_{\bm{\psi}_2}$, $\phi_{\bm{\delta}}$: networks for policy, discriminator, Q functions and encoder. \\
$T_{\text{train}}$, $\sigma(t)$, $d$, aug, $c$, $\tau$, $B$, $\alpha$, $\alpha_D$, $\gamma$, $\eta$: training steps, scheduled standard deviation, frames stack dimension, stochastic data augmentation,  clip value, target update rate, batch size, learning rate, discriminator learning rate, discount factor and temperature parameter. \\
\For{$t=1, \dots, T_{\text{\normalfont{train}}}$}{
$\sigma_t \leftarrow \sigma(t)$ \\
\If{$t \geq d-1$}{$z_t \leftarrow \phi_{\bm{\delta}}(x^T_{t-d+1:t})$}
\Else{$z_t \leftarrow \phi_{\bm{\delta}}(x^T_{0:t})$} 
$a_t \leftarrow \pi_{\bm{\theta}}(z_t) + \epsilon$ and $\epsilon \sim \mathcal{N}(0,\sigma_t^2)$ \\
$s_{t+1} \sim \mathcal{T}(\cdot|s_t, a_t)$ and $x_{t+1} \sim \mathcal{U}_T(\cdot|s_{t+1})$ \\
$\mathcal{B} \leftarrow \mathcal{B} \cup (x^T_t, a_t, x^T_{t+1})$ \\
UpdateEncoder($\mathcal{B}$) \\
UpdateDiscriminator($\mathcal{B}$, $\mathcal{B}_E$) \\
UpdateCritic($\mathcal{B}$) \\
UpdateActor($\mathcal{B}$)\\
}
\Begin(UpdateEncoder){
$\{(x^T_{t-d+1:t}, a_t, x^T_{t-d+2:t+1})\} \sim \mathcal{B}$ (sample $B$ transitions)\\ 
$\tilde z_{\bm{\delta}}(i) \leftarrow \phi_{\bm{\delta}}(\text{aug}(x^T_{t-d+1:t})_i)$ and $\tilde z_{\bm{\delta}}(j)\leftarrow \phi_{\bm{\delta}}(\text{aug}(x^T_{t-d+1:t})_j)$ \\
Update $\phi_{\bm{\delta}}$ to minimize \eqref{eq:contr_loss} with learning rate $\alpha$
}
\Begin(UpdateDiscriminator){
$\{(x^S_{t-d+1:t}, x^S_{t-d+2:t+1})\} \sim \mathcal{B}_{E}$ and $\{(x^T_{t-d+1:t}, x^T_{t-d+2:t+1})\} \sim \mathcal{B}$  (sample $B$ transitions)\\
$z_{\bm{\delta}} \leftarrow \phi_{\bm{\delta}}(x_{t-d+1:t})$ and $z_{\bm{\delta}}' \leftarrow \phi_{\bm{\delta}}(x_{t-d+2:t+1})$ for both agent and expert \\
Update $D_{\bm{\chi}}$ to minimize BCE loss with learning rate $\alpha_D$
}
\Begin(UpdateCritic){
$\{(x^T_{t-d+1:t}, a_t, x^T_{t-d+2:t+1})\} \sim \mathcal{B}$ (sample $B$ transitions)\\ 
$\tilde z_{\bm{\delta}} \leftarrow \phi_{\bm{\delta}}(\text{aug}(x^T_{t-d+1:t}))$ and $\tilde z_{\bm{\delta}}' \leftarrow \phi_{\bm{\delta}}(\text{aug}(x^T_{t-d+2:t+1}))$ \\
$a_{t+1} \leftarrow $ $\pi_{\bm{\theta}}(\tilde z_{\bm{\delta}}') + \epsilon$ and $\epsilon \sim \text{clip}(\mathcal{N}(0,\sigma_t^2), -c, c)$ \\
Update $Q_{\bm{\psi}_1}$, $Q_{\bm{\psi}_2}$ and $\phi_{\bm{\delta}}$ to minimize \eqref{eq:Q_regression_regularized} with $r_{\bm{\chi}}(z_{\bm{\delta}}, z_{\bm{\delta}}')$ and learning rate $\alpha$ \\
$\bar{\bm{\psi}}_k \leftarrow (1 - \tau)\bar{\bm{\psi}}_k +\tau\bm{\psi}_k \ \ \ \forall k \in \{1,2\}$
}
\Begin(UpdateActor){
$\{x^T_{t-d+1:t}\} \sim \mathcal{B}$ (sample $B$ observations)\\
$\tilde z_{\bm{\delta}} \leftarrow \phi_{\bm{\delta}}(\text{aug}(x^T_{t-d+1:t}))$ \\ 
$a_t \leftarrow \pi_{\bm{\theta}}(\tilde z_{\bm{\delta}}) + \epsilon$ and $\epsilon \sim \text{clip}(\mathcal{N}(0,\sigma_t^2), -c, c)$ \\
Update $\pi_{\bm{\theta}}$ using DDPG~\cite{lillicrap2015continuous} with learning rate $\alpha$
}
\end{algorithm}

\begin{table}[h!]
\centering
\caption{Hyperparameter values for C-LAIfO experiments.}
\label{tab:Hyper_1}
\small
\begin{tabular}{c c c c}\toprule
\multicolumn{2}{l}{Hyperparameter Name} & \multicolumn{2}{c}{Value}\\
\cmidrule(lr){1-2} \cmidrule(lr){3-4}
\multicolumn{2}{l}{Frames stack $(d)$} & \multicolumn{2}{c}{$3$} \\
\multicolumn{2}{l}{Discount factor $(\gamma)$} & \multicolumn{2}{c}{$0.99$} \\
\multicolumn{2}{l}{Image size} & \multicolumn{2}{c}{$64 \times 64$} \\
\multicolumn{2}{l}{Batch size $(B)$} & \multicolumn{2}{c}{$256$} \\
\multicolumn{2}{l}{Optimizer} & \multicolumn{2}{c}{Adam}\\
\multicolumn{2}{l}{Learning rate $(\alpha)$} & \multicolumn{2}{c}{$10^{-4}$}\\
\multicolumn{2}{l}{Discriminator learning rate $(\alpha_D)$} & \multicolumn{2}{c}{$4 \times 10^{-4}$}\\
\multicolumn{2}{l}{Target update rate $(\tau)$} & \multicolumn{2}{c}{$0.01$}\\
\multicolumn{2}{l}{Clip value $(c)$} & \multicolumn{2}{c}{$0.3$}\\
\multicolumn{2}{l}{Temperature parameter $(\eta)$} & \multicolumn{2}{c}{$1.0$}\\
\bottomrule
\end{tabular}
\end{table}
\newpage

\subsection{Data augmentation}
\begin{figure*}[h!]
    \centering
    \begin{subfigure}[t]{\linewidth}
        \centering
        \includegraphics[width=0.7\linewidth]{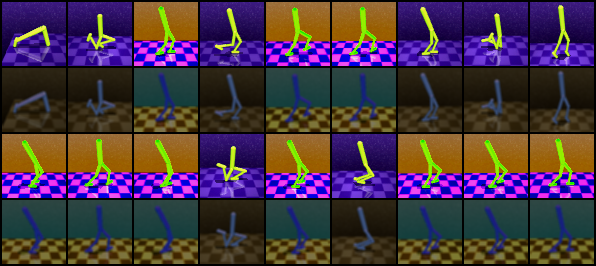}
    \end{subfigure}
    ~
    \begin{subfigure}[t]{\linewidth}
        \centering
        \includegraphics[width=0.7\linewidth]{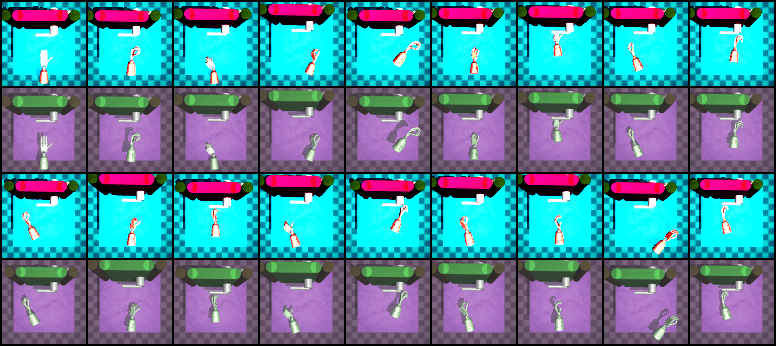}
    \end{subfigure}
    \caption{Examples of augmentation as color transformation.}
    \label{fig_app:color_aug}
\end{figure*}

\begin{figure*}[h!]
    \centering
    \begin{subfigure}[t]{\linewidth}
        \centering
        \includegraphics[width=0.7\linewidth]{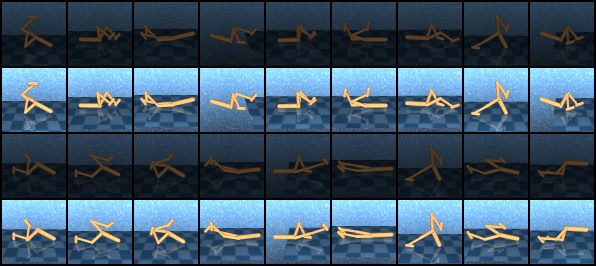}
    \end{subfigure}
    ~
    \begin{subfigure}[t]{\linewidth}
        \centering
        \includegraphics[width=0.7\linewidth]{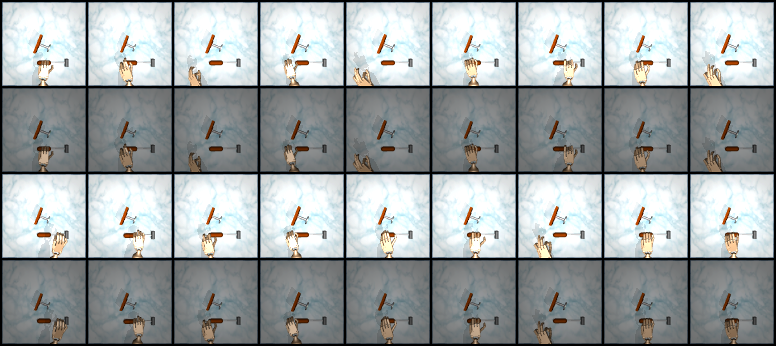}
    \end{subfigure}
    \caption{Examples of augmentation as brightness transformation.}
    \label{fig_app:light_aug}
\end{figure*}

The operations used in our data augmentation functions for the experiments in Section~\ref{sec:experiments} are summarized in Table~\ref{table:data_aug_ops}. Fig.~\ref{fig_app:color_aug} shows an example of color augmentation as implemented for the experiments in Table~\ref{table_visual_experiments}, where we randomly perform all the operations in the color transformations row in Table~\ref{table:data_aug_ops}. On the other hand, Fig.~\ref{fig_app:light_aug} shows only a brightness transformation as implemented in the ablation study in Table.~\ref{table_ablation}. Full augmentation in Table~\ref{table_ablation} performs all the operations in Table~\ref{table:data_aug_ops}. For additional details refer to our \href{https://github.com/VittorioGiammarino/C-LAIfO}{code}\footnote{https://github.com/VittorioGiammarino/C-LAIfO}.

\begin{table}[h!]
    \centering
    \caption{Operations used in our data augmentation function.}
    \label{table:data_aug_ops}
    \begin{tabular}{c|c} \toprule
         &  Operations \\
         \cmidrule(lr){1-2}
         \multirow{7}{*}{\makecell{Color \\ transformations}} & Brightness \\
         & Contrast \\
         & Saturation \\
         & Hue \\
         & Grayscale \\
         & Gaussian blur \\
         & Invert \\
         \cmidrule(lr){1-2}
         \multirow{3}{*}{\makecell{Affine \\ transformations}} & Horizontal flip \\
         & Vertical flip \\
         & Resized crop \\
         \bottomrule
    \end{tabular}

\end{table}

\label{sec_app:data_aug}

\subsection{Learning curves}
\label{sec:app_add_experiments}

All the experiments are run using Nvidia-A40 GPUs on an internal cluster. For each algorithm, we run two experiments in parallel on the same GPU and each experiment takes 1 to 4 days depending on the simulated environment and the considered algorithm. For all the implementation details refer to our \href{https://anonymous.4open.science/r/C-LAIfO-44E5/README.md}{code}. 

\begin{figure*}
    \centering
    \begin{subfigure}[t]{0.2\linewidth}
        \centering
        \includegraphics[width=\linewidth]{Figures/walker_walk_delta=0.20.png}
        \caption{}
        \label{fig_app:walker_walk_Delta=0.20}
    \end{subfigure}
    ~
    \begin{subfigure}[t]{0.2\linewidth}
        \centering
        \includegraphics[width=\linewidth]{Figures/walker_walk_delta=0.15.png}
        \caption{}
        \label{fig_app:walker_walk_Delta=0.15}
    \end{subfigure}
    ~
    \begin{subfigure}[t]{0.2\linewidth}
        \centering
        \includegraphics[width=\linewidth]{Figures/walker_walk_delta=-0.25.png}
        \caption{}
        \label{fig_app:walker_walk_Delta=-0.25}
    \end{subfigure}
    \includegraphics[width=\textwidth]{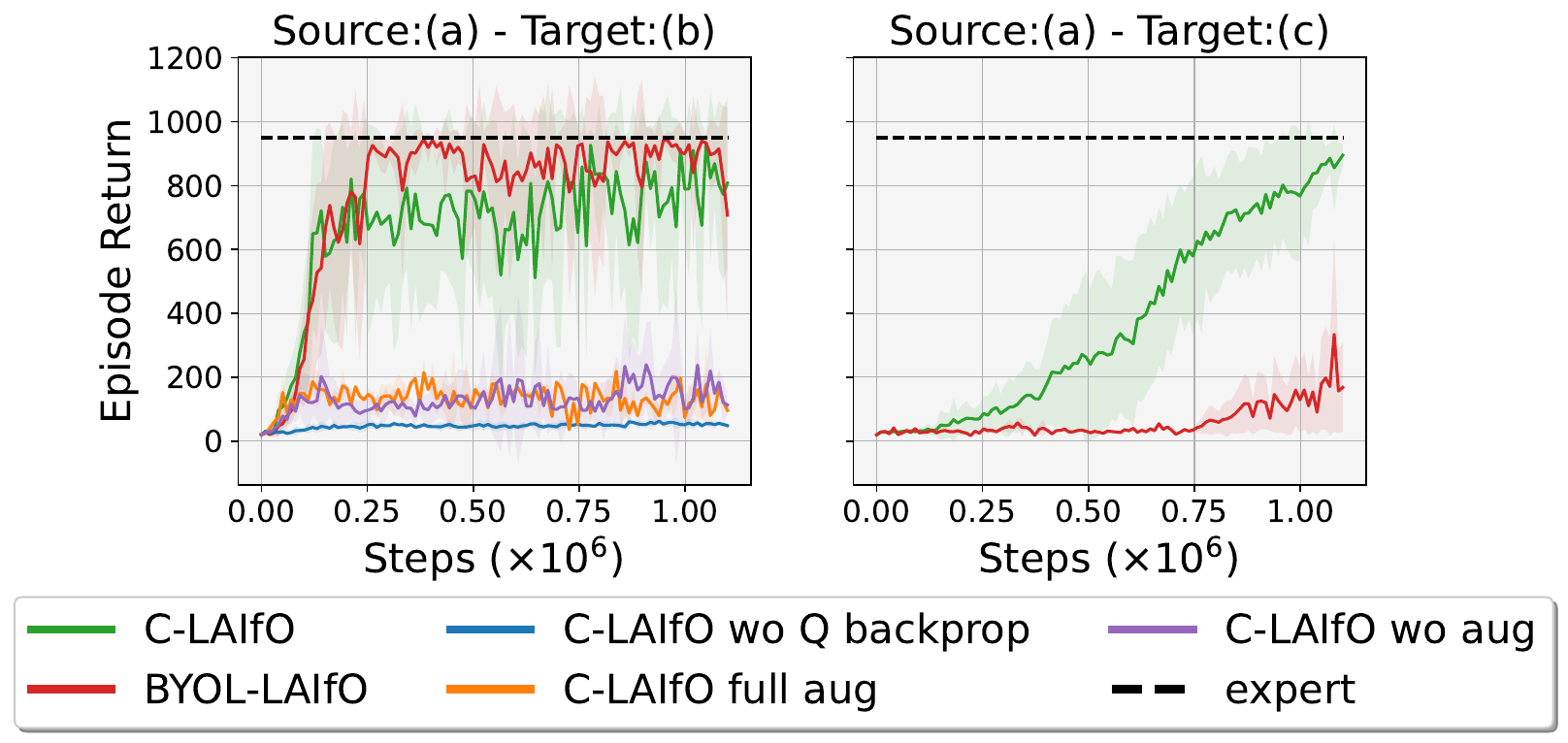}
    \caption{Learning curves for the results in Table~\ref{table_ablation}. Plots show the average return per episode as a function of training steps. The environment in \eqref{fig_app:walker_walk_Delta=0.20} represents the source-POMDP used to collect expert data, while \eqref{fig_app:walker_walk_Delta=0.15}--\eqref{fig_app:walker_walk_Delta=-0.25} are different target-POMDPs. In these experiments, visual mismatch is introduced by varying the light intensity in the target-POMDPs.}
    \label{fig_app:ablation_exp}
\end{figure*}

\begin{figure*}
    \centering
    \includegraphics[width=0.5\linewidth]{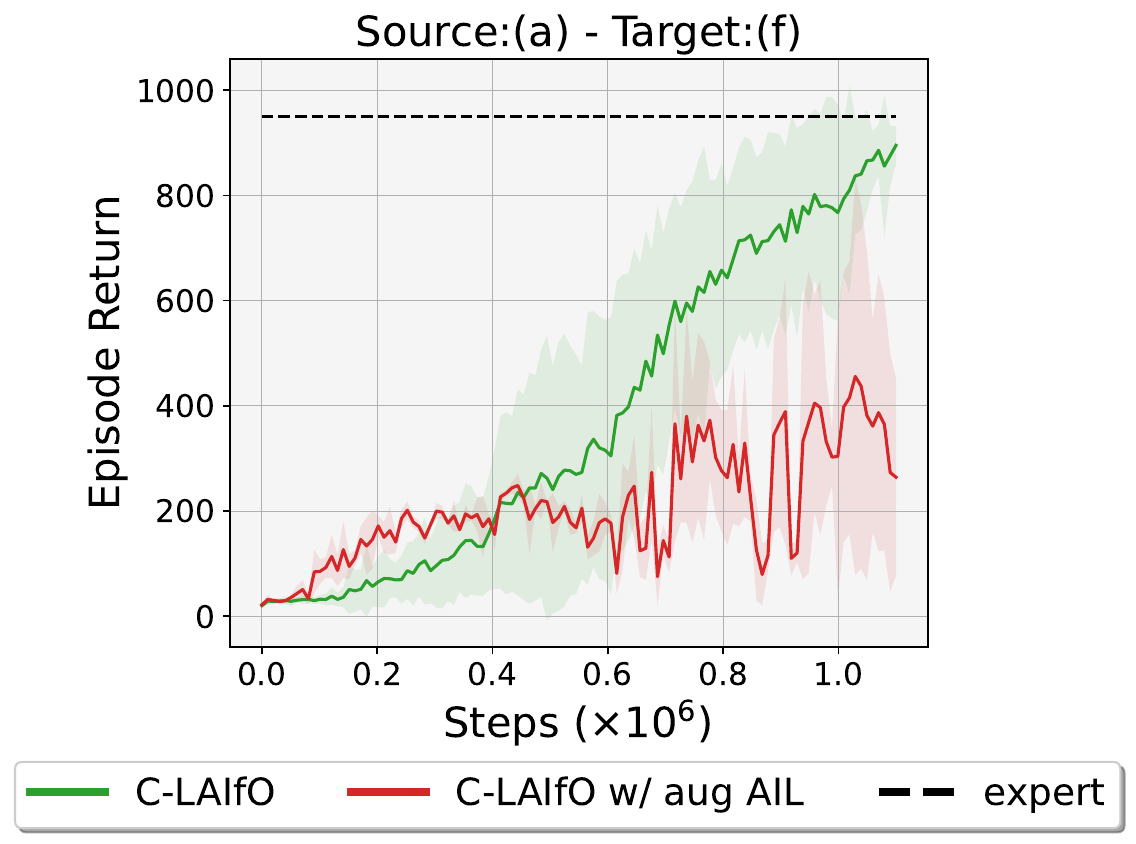}
    \caption{Experiments on the use of data augmentation in AIL in \eqref{eq:AIL_BCE}. Plots show the average return per episode as a function of training steps. The environment in \eqref{fig_app:walker_walk_Delta=0.20} represents the source-POMDP used to collect expert data, while \eqref{fig_app:walker_walk_Delta=-0.25} the target-POMDPs. These experiments show how randomized data augmentation used during AIL triggers instability.}
    \label{fig_app:aug_effect_on_AIL}
\end{figure*}

\begin{figure*}
    \centering
    \begin{subfigure}[t]{0.2\linewidth}
        \centering
        \includegraphics[width=\linewidth]{Figures/walker_walk_delta=-0.25.png}
        \caption{}
        \label{fig_app:walker_dark}
    \end{subfigure}
    ~
    \begin{subfigure}[t]{0.2\linewidth}
        \centering
        \includegraphics[width=\linewidth]{Figures/walker_walk_delta=0.20.png}
        \caption{}
        \label{fig_app:walker_standard}
    \end{subfigure}
    ~
    \begin{subfigure}[t]{0.2\linewidth}
        \centering
        \includegraphics[width=\linewidth]{Figures/Source_all_tg.png}
        \caption{}
        \label{fig_app:walker_full}
    \end{subfigure}
    ~
    \includegraphics[width=\textwidth]{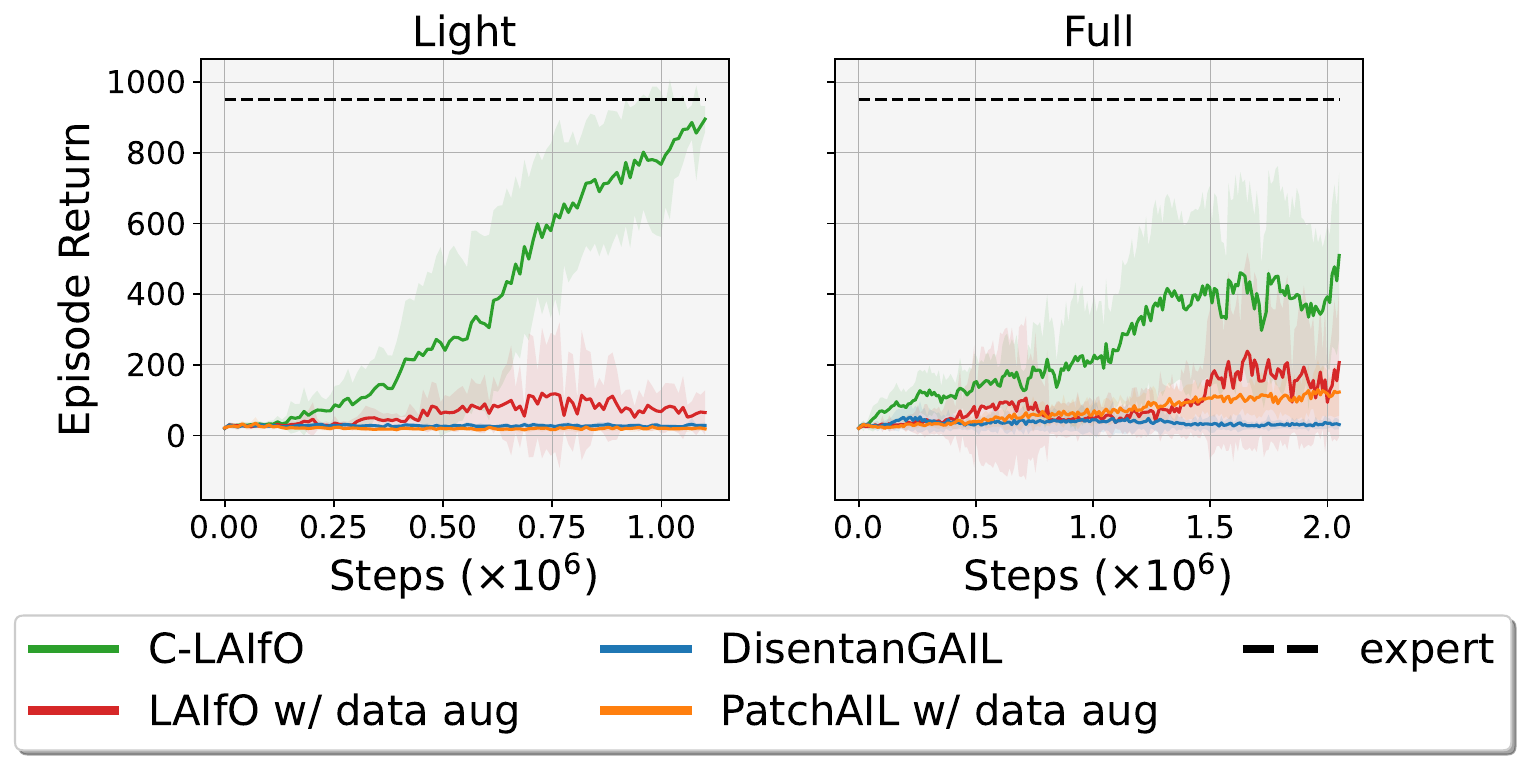}
    \caption{Learning curves for the results in Table~\ref{table_visual_experiments}. Plots show the average return per episode as a function of training steps. The Light experiment consists in \eqref{fig_app:walker_standard} as source-POMDP and \eqref{fig_app:walker_dark} as target-POMDP. The Full experiment considers \eqref{fig_app:walker_standard} as target-POMDP and \eqref{fig_app:walker_full} as source-POMDP.}
    \label{fig_app:visual_exp}
\end{figure*}

\begin{figure*}
    \centering
        \begin{subfigure}[t]{0.09\linewidth}
        \centering
        \includegraphics[width=\linewidth]{Figures/Door.png}
        \caption{}
        \label{fig_app:door}
    \end{subfigure}
    ~
    \begin{subfigure}[t]{0.09\linewidth}
        \centering
        \includegraphics[width=\linewidth]{Figures/door_mismatch.png}
        \caption{}
        \label{fig_app:door-light}
    \end{subfigure}
    ~
    \begin{subfigure}[t]{0.09\linewidth}
        \centering
        \includegraphics[width=\linewidth]{Figures/Door_color.png}
        \caption{}
        \label{fig_app:door-color}
    \end{subfigure}
    ~
    \begin{subfigure}[t]{0.09\linewidth}
        \centering
        \includegraphics[width=\linewidth]{Figures/hammer.png}
        \caption{}
        \label{fig_app:hammer}
    \end{subfigure}
    ~
    \begin{subfigure}[t]{0.09\linewidth}
        \centering
        \includegraphics[width=\linewidth]{Figures/hammer_mismatch.png}
        \caption{}
        \label{fig_app:hammer-light}
    \end{subfigure}
    ~
    \begin{subfigure}[t]{0.09\linewidth}
        \centering
        \includegraphics[width=\linewidth]{Figures/hammer_color.png}
        \caption{}
        \label{fig_app:hammer-color}
    \end{subfigure}
    ~
    \begin{subfigure}[t]{0.09\linewidth}
        \centering
        \includegraphics[width=\linewidth]{Figures/pen.png}
        \caption{}
        \label{fig_app:pen}
    \end{subfigure}
    ~
    \begin{subfigure}[t]{0.09\linewidth}
        \centering
        \includegraphics[width=\linewidth]{Figures/pen_mismatch.png}
        \caption{}
        \label{fig_app:pen-light}
    \end{subfigure}
    ~
    \begin{subfigure}[t]{0.09\linewidth}
        \centering
        \includegraphics[width=\linewidth]{Figures/pen_color.png}
        \caption{}
        \label{fig_app:pen-color}
    \end{subfigure}
    \includegraphics[width=0.7\textwidth]{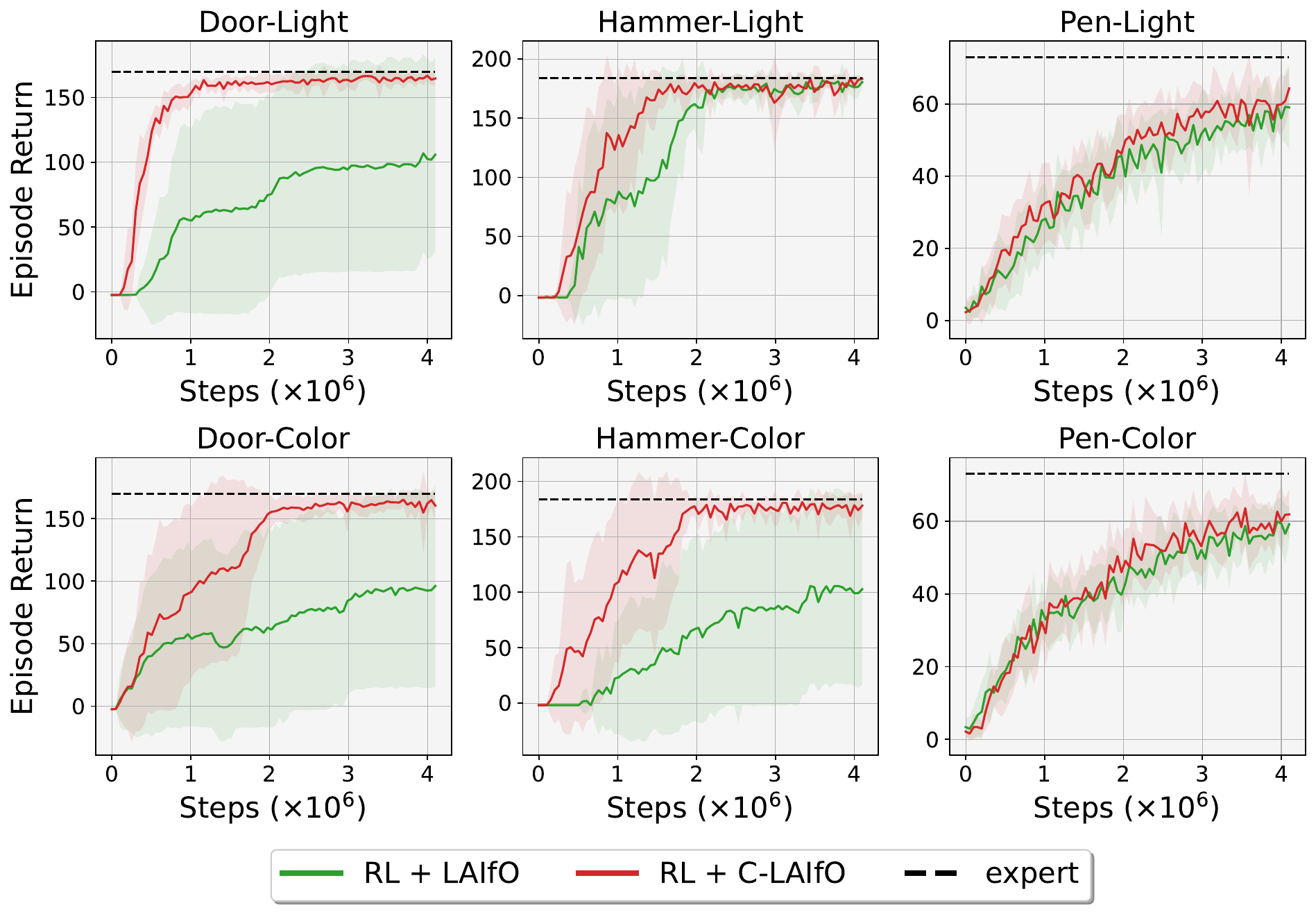}
    \caption{Learning curves for the results in Table~\ref{table_adroit}. Plots show the average return per episode as a function of training steps. The Door-Light and Door-Color experiments consider \eqref{fig_app:door} as the source-POMDP and \eqref{fig_app:door-light} and \eqref{fig_app:door-color} as the respective target-POMDPs. Similarly, the Hammer-Light and Hammer-Color experiments consider \eqref{fig_app:hammer} as source-POMDP and \eqref{fig_app:hammer-light} and \eqref{fig_app:hammer-color} as the respective target-POMDPs, while the Pen-Light and Pen-Color experiments consider \eqref{fig_app:pen} as source-POMDP and \eqref{fig_app:pen-light} and \eqref{fig_app:pen-color} as the respective target-POMDPs.}
    \label{fig_app:adroit_experiments}
\end{figure*}

Fig.~\ref{fig_app:ablation_exp}, Fig.~\ref{fig_app:visual_exp}, and Fig.~\ref{fig_app:adroit_experiments} show the learning curves for the results in Table~\ref{table_ablation}, Table~\ref{table_visual_experiments}, and Table~\ref{table_adroit}, respectively. Fig.~\ref{fig_app:aug_effect_on_AIL} shows the effect of randomized data augmentation when used in the AIL step in \eqref{eq:AIL_BCE}.

\subsection{PCA and t-SNE analysis}
\label{sec_app:pca_t-sne}

\begin{figure*}[h!]
    \centering
    \begin{subfigure}[t]{0.2\linewidth}
        \centering
        \includegraphics[width=0.5\linewidth]{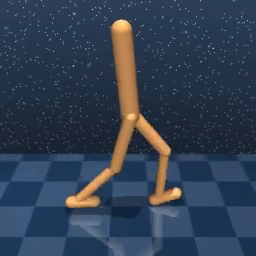}
        \caption{Source.}
        \label{fig_app:source_optimal_light}
    \end{subfigure}
    ~
    \begin{subfigure}[t]{0.2\linewidth}
        \centering
        \includegraphics[width=0.5\linewidth]{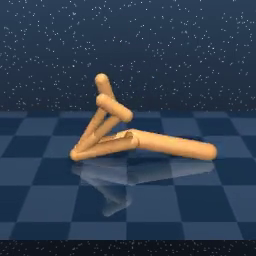}
        \caption{Source Random.}
        \label{fig_app:source_random_light}
    \end{subfigure}
    ~
    \begin{subfigure}[t]{0.2\linewidth}
        \centering
        \includegraphics[width=0.5\linewidth]{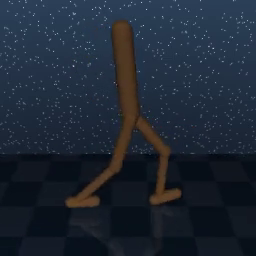}
        \caption{Target.}
        \label{fig_app:target_optimal_light}
    \end{subfigure}
    ~
    \begin{subfigure}[t]{0.2\linewidth}
        \centering
        \includegraphics[width=0.5\linewidth]{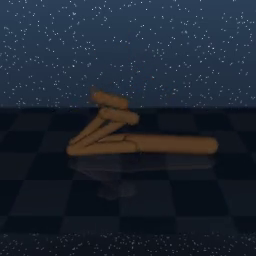}
        \caption{Target Random.}
        \label{fig_app:target_random_light}
    \end{subfigure}
    ~
    \begin{subfigure}[t]{\linewidth}
        \centering
        \includegraphics[width=\linewidth]{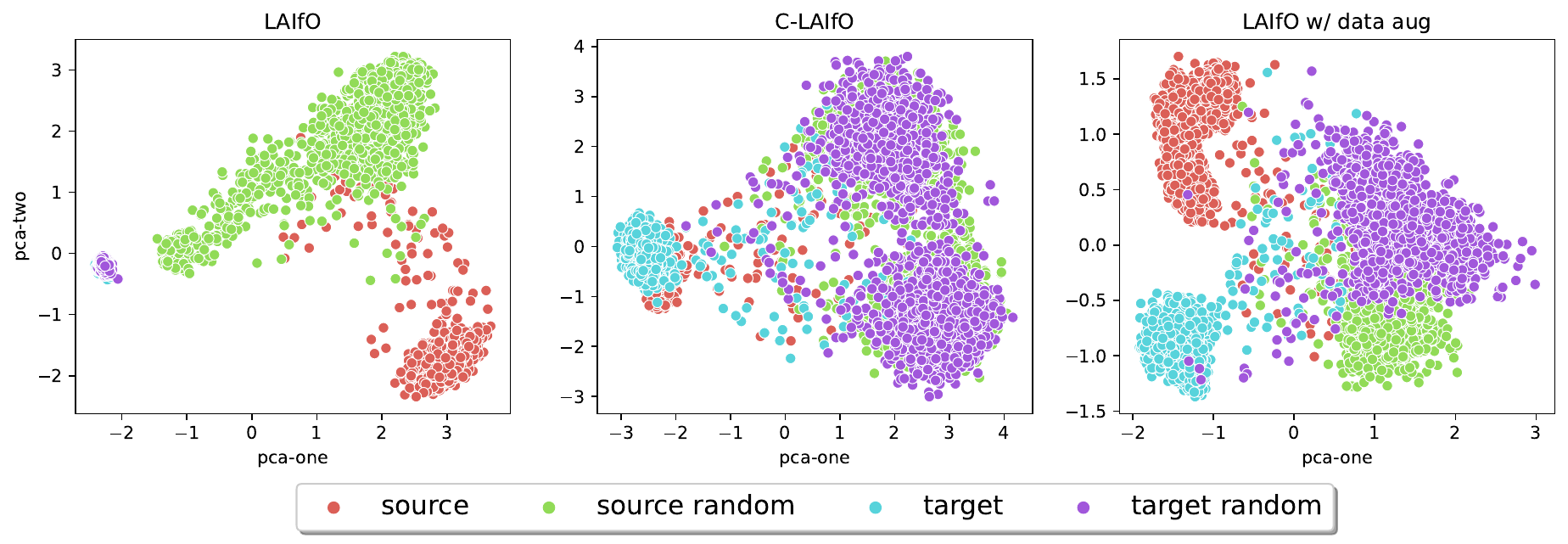}
        \label{fig_app:PCA_light}
    \end{subfigure}
    ~
    \begin{subfigure}[t]{\linewidth}
        \centering
        \includegraphics[width=\linewidth]{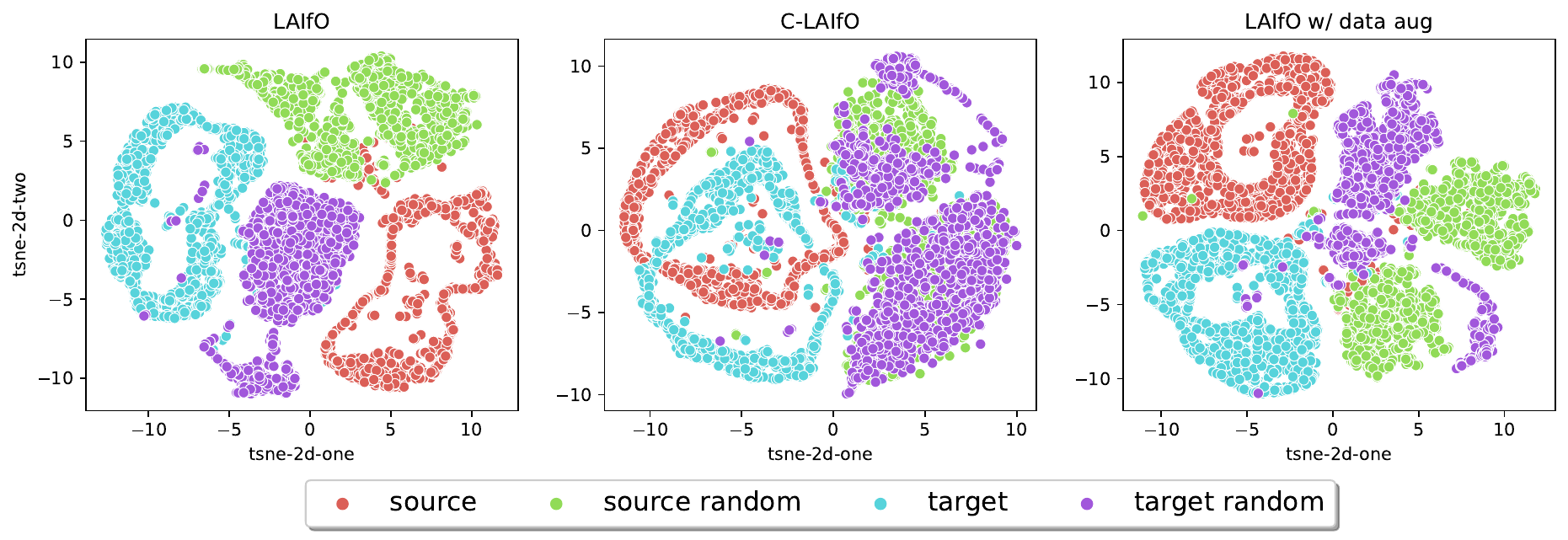}
        \label{fig_app:t-SNE_light}
    \end{subfigure}
    \caption{PCA and t-SNE visualizations for the Light experiment in Table~\ref{table_visual_experiments}.}
    \label{fig_app:walker_walk_light_mismatch}
\end{figure*}

\begin{figure*}[h!]
    \centering
    \begin{subfigure}[t]{0.2\linewidth}
        \centering
        \captionsetup{justification=centering}
        \includegraphics[width=0.5\linewidth]{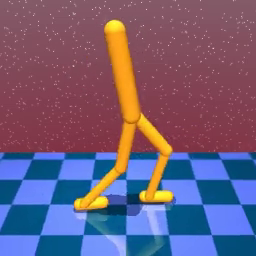}
        \caption{Source\\(Full).}
        \label{fig_app:source_optimal_color}
    \end{subfigure}
    ~
    \begin{subfigure}[t]{0.2\linewidth}
        \centering
        \captionsetup{justification=centering}
        \includegraphics[width=0.5\linewidth]{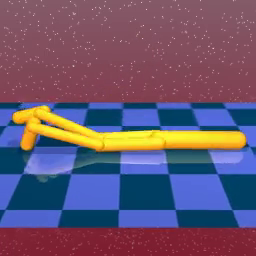}
        \caption{Source Random\\(Full).}
        \label{fig_app:source_random_color}
    \end{subfigure}
    ~
    \begin{subfigure}[t]{0.2\linewidth}
        \centering
        \captionsetup{justification=centering}
        \includegraphics[width=0.5\linewidth]{Figures/source_optimal_colors_unseen.png}
        \caption{Source\\(Full-unseen).}
        \label{fig_app:source_optimal_color_unseen}
    \end{subfigure}
    ~
    \begin{subfigure}[t]{0.2\linewidth}
        \centering
        \captionsetup{justification=centering}
        \includegraphics[width=0.5\linewidth]{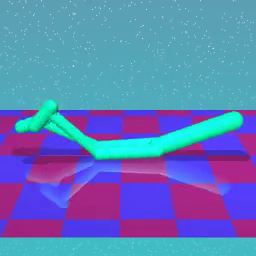}
        \caption{Source Random\\(Full-unseen).}
        \label{fig_app:source_random_color_unseen}
    \end{subfigure}
    ~
    \begin{subfigure}[t]{0.2\linewidth}
        \centering
        \includegraphics[width=0.5\linewidth]{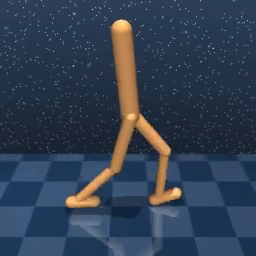}
        \caption{Target.}
        \label{fig_app:target_optimal}
    \end{subfigure}
    ~
    \begin{subfigure}[t]{0.2\linewidth}
        \centering
        \includegraphics[width=0.5\linewidth]{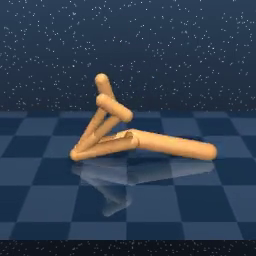}
        \caption{Target Random.}
        \label{fig_app:target_random}
    \end{subfigure}
    ~
    \begin{subfigure}[t]{0.7\linewidth}
        \centering
        \includegraphics[width=\linewidth]{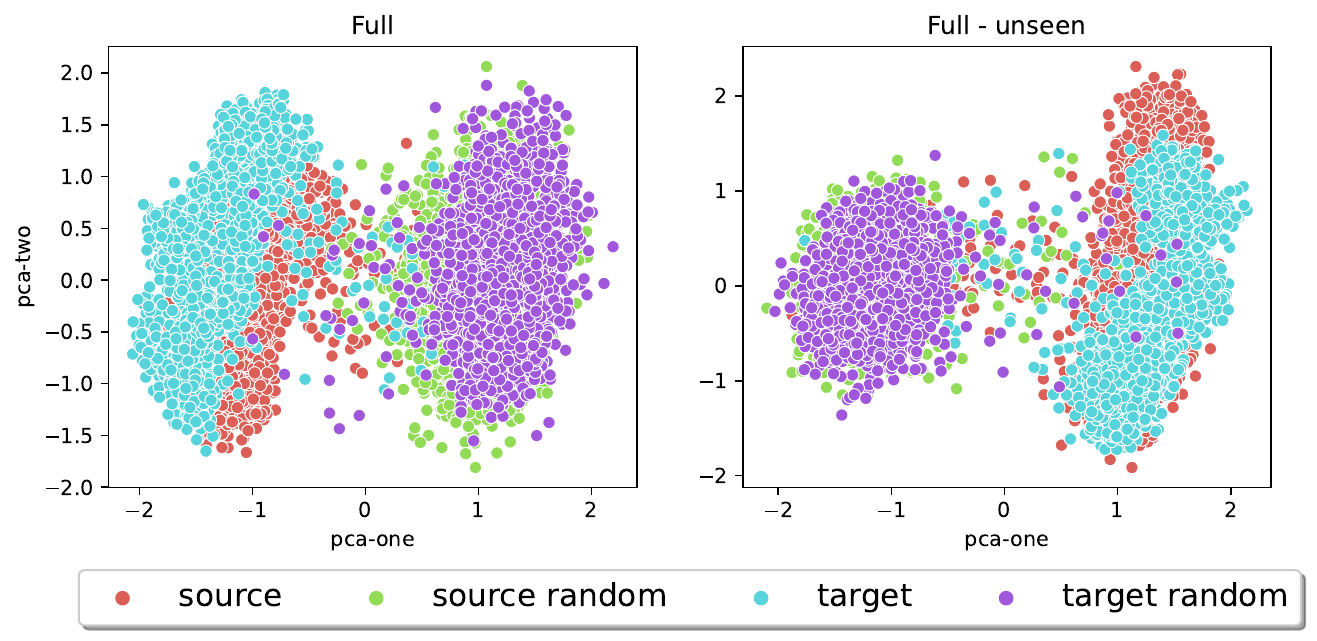}
        \label{fig_app:PCA_color}
    \end{subfigure}
    ~
    \begin{subfigure}[t]{0.7\linewidth}
        \centering
        \includegraphics[width=\linewidth]{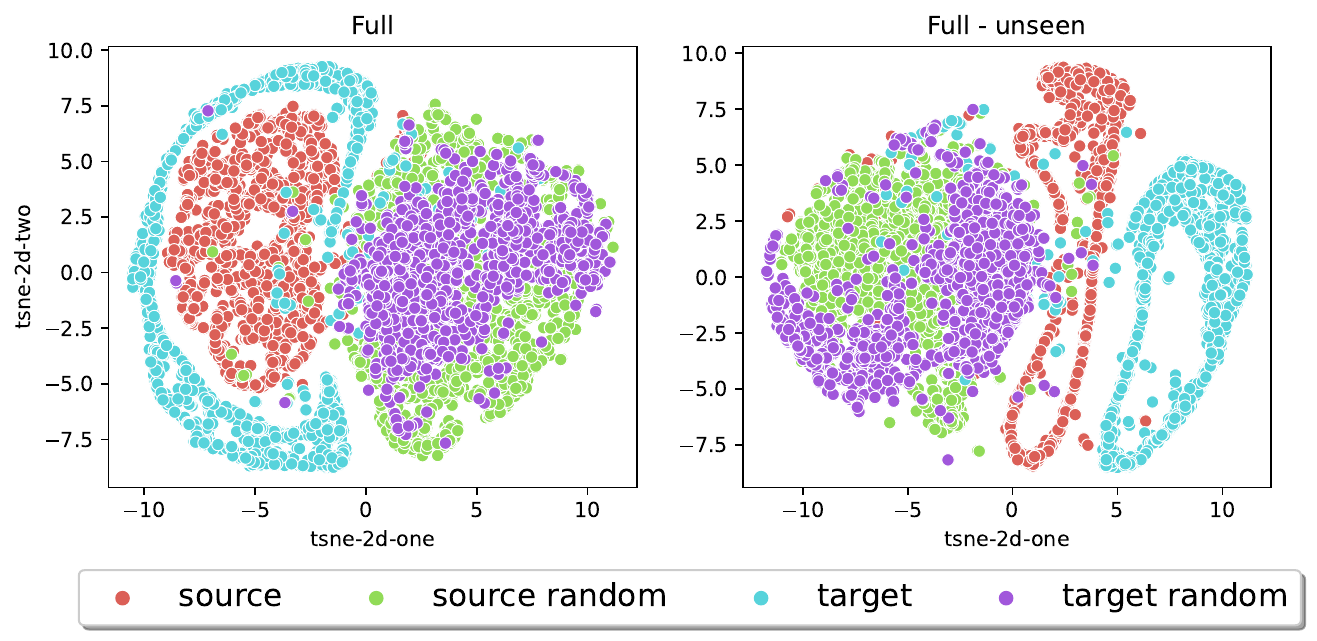}
        \label{fig_app:t-SNE_color}
    \end{subfigure}
    \caption{PCA and t-SNE for the Full experiment in Table~\ref{table_visual_experiments}. Fig.~\ref{fig_app:source_optimal_color} and Fig.~\ref{fig_app:source_random_color} denote the source environment for the Full experiment (left figures in Fig.~\ref{fig_app:walker_walk_color_mismatch}). Fig.~\ref{fig_app:source_optimal_color_unseen} and Fig.~\ref{fig_app:source_random_color_unseen} denote the source environment for the Full-unseen experiment (right figures in Fig.~\ref{fig_app:walker_walk_color_mismatch}). }
    \label{fig_app:walker_walk_color_mismatch}
\end{figure*}

In Fig.~\ref{fig_app:walker_walk_light_mismatch} we compare the latent space $\mathcal{Z}$ learned by LAIfO, C-LAIfO, and LAIfO with data augmentation in the Light setting from Table~\ref{table_visual_experiments} and Fig.~\ref{fig_app:ablation_exp}. Specifically, we define \textit{source} and \textit{target} as the observations generated by an optimal policy in the source and target POMDPs, respectively. Similarly, \textit{source random} and \textit{target random} are observations generated by random policies. These observations are processed with the encoder $\phi_{\bm{\delta}}:\mathcal{X}^d \to \mathcal{Z}$ trained using the respective algorithms. We perform PCA and t-SNE on this set of latent variables $z_{1:T}$ and plot the first two principal components. The results show that C-LAIfO is the only algorithm capable of filtering out visual distractors and clustering together data points with the same goal-completion information. In Fig.~\ref{fig_app:walker_walk_color_mismatch}, we focus on C-LAIfO and test for generalization to the unseen environment in Fig.~\ref{fig_app:source_optimal_color_unseen}. We train $\phi_{\bm{\delta}}$ on the Full setting from Table~\ref{table_visual_experiments} and Fig.~\ref{fig_app:visual_exp} and perform PCA and t-SNE as described. In Fig.~\ref{fig_app:walker_walk_color_mismatch}, the Full figures have \eqref{fig_app:source_optimal_color}-\eqref{fig_app:source_random_color} as source-POMDP and \eqref{fig_app:target_optimal}-\eqref{fig_app:target_random} as target-POMDP. Similarly the Full-unseen figures have \eqref{fig_app:source_optimal_color_unseen}-\eqref{fig_app:source_random_color_unseen} as source-POMDP and \eqref{fig_app:target_optimal}-\eqref{fig_app:target_random} as target-POMDP. The results in the Full-unseen setting match those in the Full experiment, indicating that $\phi_{\bm{\delta}}$ trained on \eqref{fig_app:source_optimal_color} can successfully generalize to \eqref{fig_app:source_optimal_color_unseen}.

\end{document}